\documentclass[lettersize,journal]{IEEEtran}

\usepackage{amsmath,amsfonts}
\usepackage{algorithmicx}
\usepackage[ruled]{algorithm2e}
\usepackage{array}
\usepackage[caption=false,font=normalsize,labelfont=sf,textfont=sf]{subfig}
\usepackage{textcomp}
\usepackage{stfloats}
\usepackage{url}
\usepackage{verbatim}
\usepackage{graphicx}
\usepackage{cite}
\usepackage{cases}
\usepackage{bm}
\usepackage{amssymb}
\usepackage{diagbox}
\usepackage{multirow}
\usepackage{epstopdf}
\usepackage{epsfig}
 \usepackage{amsthm}
 \usepackage{color}
 \usepackage{xcolor}
\usepackage{adjustbox}
\usepackage{fancyhdr}

\begin{document}
\title{LUIEO: A Lightweight Model for Integrating Underwater Image Enhancement and Object Detection}

\author{Bin Li,
        Li Li,
        Zhenwei Zhang$^*$,
        Yuping Duan
\thanks{B. Li, Z. Zhang  are with School of Mathematics, North University of China, Taiyuan, Shanxi, 030051, China.
E-mail: 20230071@nuc.edu.cn. \emph{Asterisk indicates the corresponding author.}}
\thanks{L.Li is with College of Computer and Information Technology, Shanxi University, Taiyuan, Shanxi, 030051, China. }
\thanks{Y.Duan is with School of Mathematics Sciences, Beijing Normal University, Beijing, 10091, China. }
\thanks{Manuscript received April 19, 2021; revised August 16, 2021.}
}

\markboth{Journal of \LaTeX\ Class Files,~Vol.~14, No.~8, August~2021}%
{Shell \MakeLowercase{\textit{et al.}}: A Sample Article Using IEEEtran.cls for IEEE Journals}

\IEEEpubid{0000--0000/00\$00.00~\copyright~2021 IEEE}


\maketitle

\renewcommand{\headrulewidth}{0mm}
\begin{abstract}
Underwater optical images inevitably suffer from various degradation factors such as blurring, low contrast, and color distortion, which hinder the accuracy of object detection tasks.
Due to the lack of paired underwater/clean images, most research methods adopt a strategy of first enhancing and then detecting,  resulting in a lack of feature communication between the two learning tasks.
On the other hand, due to the contradiction between the diverse degradation factors of underwater images and the limited number of samples, existing underwater enhancement methods are difficult to effectively enhance degraded images of unknown water bodies, thereby limiting the improvement of object detection accuracy.
Therefore, most underwater target detection results are still displayed on degraded images, making it difficult to visually judge the correctness of the detection results.
To address the above issues, this paper proposes a multi-task learning method that simultaneously enhances underwater images and improves detection accuracy.
Compared with single-task learning, the integrated model allows for the dynamic adjustment of information communication and sharing between different tasks.
For image enhancement tasks, this article uses refined simulation formulas to provide prior information and physical constraints to the model, which effectively improves the model's generalization ability.
Therefore, this article introduces a physical module to decompose underwater images into clean images, background light, and transmission images and uses a physical model to calculate underwater images for self-supervision.
Due to the fact that real underwater images can only provide annotated object labels, this paper introduces physical constraints to ensure that object detection tasks do not interfere with image enhancement tasks.
Numerical experiments demonstrate that the proposed model achieves satisfactory results in visual performance, object detection accuracy, and detection efficiency compared to state-of-the-art comparative methods.
All our codes and data are available at \url{https://github.com/DrZhangZW/LUIEO}.
\end{abstract}

\begin{IEEEkeywords}
 Image enhancement; Underwater object detection; Lightweight
\end{IEEEkeywords}

\section{Introduction}

Underwater object detection has significant application value in marine monitoring, underwater resource exploration, intelligent aquaculture, and other fields.
However, due to the absorption and scattering of light in water, underwater images suffer from issues such as noise, low contrast, and color degradation \cite{jha2024cbla}, making it difficult to achieve satisfactory detection accuracy on degraded images.
As a result, underwater object detection is a multi-task learning problem that combines image enhancement and object detection.
Limited by the computing resources of underwater vehicles, designing separate  models for image enhancement and object detection increases computing resources and inference time.
Therefore, this paper proposes a lightweight model that integrates both image enhancement and object detection tasks to complete object detection tasks.

 \IEEEpubidadjcol

Due to the inability to obtain clean underwater images, the deep learning methods for image enhancement need to address the issue of insufficient training data.
Li \emph{et al.} \cite{li2020underwater} synthesized different types of underwater images based on an underwater imaging model to enhance underwater images.
However, there remains a gap between the synthesized images and real underwater images.
To enhance the generalization of models in real underwater environments, Li \emph{et al.} \cite{li2019underwater} constructed a UIEB dataset consisting of $890$ paired images, where the reference images were selected from $12$ enhancement algorithms with the best visual performances.
However, manually selecting reference images is a time-consuming task.
Recently, some underwater datasets have been proposed to overcome the scarcity and low quality of underwater samples.
Kappor  \emph{et al.} \cite{kapoor2023domain} recreated paired underwater images by using water depth to degrade images from UIEB, and proposed an encoder-decoder network to preserve the texture and style of the images.
Peng \emph{et al.} \cite{peng2023u} built a large-scale underwater image dataset to train the U-shape Transformer network, which covers a broader range of underwater scenes and better visual reference images than existing underwater datasets.
Xie  \emph{et al.} \cite{xie2024uveb} constructed the first large-scale high-resolution underwater video enhancement benchmark to promote the development of underwater vision, and proposed the first supervised underwater video enhancement method.
Generative adversarial networks have also received significant attention in the field of underwater image enhancement.
Islam \emph{et al.} \cite{islam2020fast} utilized a CycleGAN-based method to learn
the transformation between the clean image domain and the underwater image domain, resulting in a large dataset called EUVP.
Wu \emph{et al.} \cite{wu2024self} used the imaging process of underwater scenes to reduce the amount of data required for style conversion from in-air images to underwater images, generating diverse underwater samples.
Among these methods, this article adopts the approach of using fine underwater imaging simulation to generate diverse degraded underwater images.
Compared to other methods, the simulated underwater samples follow the physical laws of underwater imaging, which can guide the network model to learn the physical process of underwater imaging.
Therefore, the physical constraints of underwater imaging provide supervised information for image enhancements task, allowing both image enhancement and object detection tasks to be trained simultaneously.


Autonomous vehicle safety driving requires many vision tasks, such as panoptic segmentation and object detection \cite{xu2023systematic,wang2024is}.
To improve detection accuracy,  object detection task often need to be coupled with image enhancement task.
The combinations of underwater enhancement and object detection can be roughly divided into preprocessing and multi-task learning methods.
Due to limited underwater computing resources, some researchers have proposed lightweight object detection models.
Yan \emph{et al.} \cite{yan2023uw} proposed a model-driven cycle-consistent generative adversarial network model to enhance underwater images, in which the enhanced images were used to detect underwater objects.
Xue \emph{et al.} \cite{xue2023investigating} proposed a multi-branch aggregation network to estimate the degradation variables of the underwater imaging model, which has been proven to improve the accuracy of underwater detection.
Cai \emph{et al.} \cite{cai2023cure} constructed a cascaded deep network to improve degraded underwater images in a coarse to fine way, in which the enhanced images effectively improve object detection results.
Zhou \emph{et al.} \cite{zhou2023lightweight} proposed a lightweight deep-water object detection network, where a lightweight attention module was used for processing to enhance underwater images.
Liu \emph{et al.} \cite{liu2024unitmodule} proposed a plug-and-play underwater joint image enhancement module that provides the input images for the detector.

Compared to independently optimizing two learning tasks, simultaneously optimizing two learning tasks can improve the ability of information exchange and sharing between different tasks.
However, due to the lack of paired clean images in the object detection datasets, there is insufficient supervised information to optimize the image enhancement model.
As a result, these methods use feature fusion or texture enhancement to assist in object detection task,  but cannot complete image enhancement task.
Zhou \emph{et al.} \cite{zhou2022yolotrashcan} proposed an efficient channel attention module and dilated parallel modules for extracting and fusing underwater targets of different scales to improve detection accuracy.
Hua \emph{et al.} \cite{hua2023underwater} designed a feature enhancement gating module to selectively suppress or enhance multi-level features, which were used to detect underwater objects by a spatial pyramid pooling structure.
Wang \emph{et al.} \cite{wang2024dual} proposed a multi-task learning method that combines image enhancement and object detection, where
the image enhancement method uses edge detection to enhance the texture information of the images.
Wang \emph{et al.} \cite{wang2023reinforcement} proposed a reinforcement learning paradigm of configuring visual enhancement for object detection in underwater scenes, where the image enhancement task serves object detection rather than human vision.
Therefore, in most methods, image enhancement is used merely as an auxiliary module to improve object detection accuracy, resulting in the detected targets
still being displayed on the degraded underwater images,
 and it is difficult to visually judge the accuracy of the detection results.

To optimize both learning tasks simultaneously,
this paper proposes a model-driven lightweight deep-learning model that integrates image enhancement and object detection.
Due to the lack of paired clean images, this paper uses a refined simulation
formula to generate various degraded underwater images, guiding the network
to learn the physical process and prior knowledge of underwater imaging.
Specifically, this paper designs a physical module to decompose underwater images into a clean image, background light, and a transmission map.
These physical variables can be used to obtain an underwater image through underwater imaging principles, providing supervised information for enhancing real underwater images.
Therefore, this self-supervised information enables image enhancement and object detection to be trained simultaneously, and optimized towards jointly improving detection accuracy and image enhancement.

For the object detection task, this paper introduces a feature pyramid network and path aggregation network on the image enhancement network to fuse multi-scale features, which improves the expressive ability of features to detect underwater targets of different sizes.
For features of different sizes, anchor-free detection heads are used to obtain the detection results, which accelerates the post-processing inference step compared to anchor-based methods.
Considering limited computing resources, this paper designs a lightweight network structure based on an inverted residual structure and the MobileViT-V3 module\cite{wadekar2022mobilevitv3}, where the inverted residual structure is a lightweight convolutional structure used to extract image features.
MobileViT-V3 is a lightweight hybrid architecture that combines CNNs and transformers, which has the advantages of CNN spatial bias induction and transformers processing of global information.
Compared to deformable transformer \cite{liu2023refined} and Swin Transformer \cite{zhou2024ad}, MobileViT-V3 is more suitable for underwater real-time tasks.
Our main contributions are summarized as follows:

\begin{enumerate}
\item To our best knowledge, this is the first lightweight model that simultaneously completes underwater image enhancement and object detection tasks, with a model size of 33.8M.
\item In this paper, a refined underwater imaging model is developed to simulate underwater images.
    Various underwater simulation images provide prior knowledge and physical guidance for enhancing the model, allowing the network to use physical constraints for self-supervised training and to train with more underwater samples.
\item The proposed model effectively enhances various degraded images and improves detection accuracy on multiple underwater datasets.
The object detection results are displayed on the enhanced images aligning with practical application scenarios.
The numerical results confirm that image enhancement tasks can improve the accuracy of object detection, increasing the mAP50 index by nearly $5.7\%$ compared to the baseline model, fully demonstrating the benefits of integrating image enhancement and object detection.
\end{enumerate}

\section{An integrated model for underwater image enhancement and object detection}\label{sec3}
The scattering and absorption of underwater suspended particles result in diverse degradation factors in underwater images, which limits the accuracy of underwater object detection.
Therefore, object detection tasks usually need to be combined with image enhancement tasks to improve the accuracy of detection.
Instead of optimizing these two subtasks independently, this paper proposes a lightweight model that integrates both image enhancement and object detection.
Given the lack of paired clean images for underwater samples, this paper uses an underwater imaging model to train a self-supervised image enhancement model.
Fig.\ref{LUIEO} shows an integrated model of image enhancement and object detection in this paper.

\subsection{Underwater image enhancement model}

Due to the absorption of seawater and the scattering of particles in water,
the signal captured by the camera is the main sum of the direct signal and the scattered signal, as shown in Fig.\ref{opmodel}.
Therefore, the underwater imaging model \cite{tan2008visibility,chiang2011underwater} can be described as follows:
\begin{equation}\label{image model}
I_\lambda(x)=J_\lambda(x)t_\lambda(x)+B_\lambda(x)(1-t_\lambda(x)),\lambda\in\{{R,G,B}\},
\end{equation}
where $\lambda\in \{R,G,B\}$ is one of the RGB channels.
Here, $I_\lambda(x)$ is the observed intensity at pixel $x$, $J_\lambda$ is the clean image,
$B_\lambda$ denotes the background light and $t_\lambda$ is the transmission map.
The transmission map $t_\lambda(x)$ is defined by $t_\lambda(x)=e^{-c_\lambda d(x)}$ with $d(x)$ being the scene distance and $c_\lambda$ being the attenuation coefficient.

The underwater imaging model \eqref{image model} can be used for the construction of simulation datasets and network structures.
As shown in Fig.\ref{LUIEO}, the underwater image enhancement network maps an underwater image  into  clean image, background light, and transmission map.
Therefore, the three physical variables predicted by the network can be used to calculate an underwater image using formula \eqref{image model}, allowing the network to perform self supervised training on real underwater images.

\subsection{Object detection model}

Compared to the strategy of first enhancing and then detecting, this paper proposes a network model that simultaneously completes image enhancement and object detection.
As shown in Fig.\ref{LUIEO}, the last layer of the encoder employs the spatial pyramid pooling fast (SPPF) module \cite{jocher2022ultralytics} to perform multi-scale pooling on the feature map to fuse features of different scales, which helps improve the performance of the model in object detection tasks.
The feature layers in the subsequent decoding process use pixel-wise addition to fuse features of the same size in the encoding layer.
This addition method does not increase the number of feature map channels and facilitates the lightweight design of the network structure.
For the object detection task, the feature layers upsampled from the decoding layer form a feature pyramid.
However, this low-resolution upsampling to a high-resolution pyramid conveys strong semantic information but lacks localization information.
Therefore, the path aggregation module is introduced to transmit localization information, by downsampling high-resolution feature maps.
These downsampled maps are then concatenated with the feature maps of the same size from the decoding layer for feature fusion.
Subsequently, the fused multi-scale features are processed by an anchor-free detection head to identify the objects and locate their bounding boxes.

\subsection{Synthetic underwater image dataset}
In the following, this paper proposes a refined simulation formula based on formula \eqref{image model} to degraded in-air images.
To simulate various underwater environments, we fully consider the interference of water types, water depths, and artificial light sources in underwater imaging.

\begin{figure}[htbp]
\centering
\includegraphics[width=0.35\textwidth]{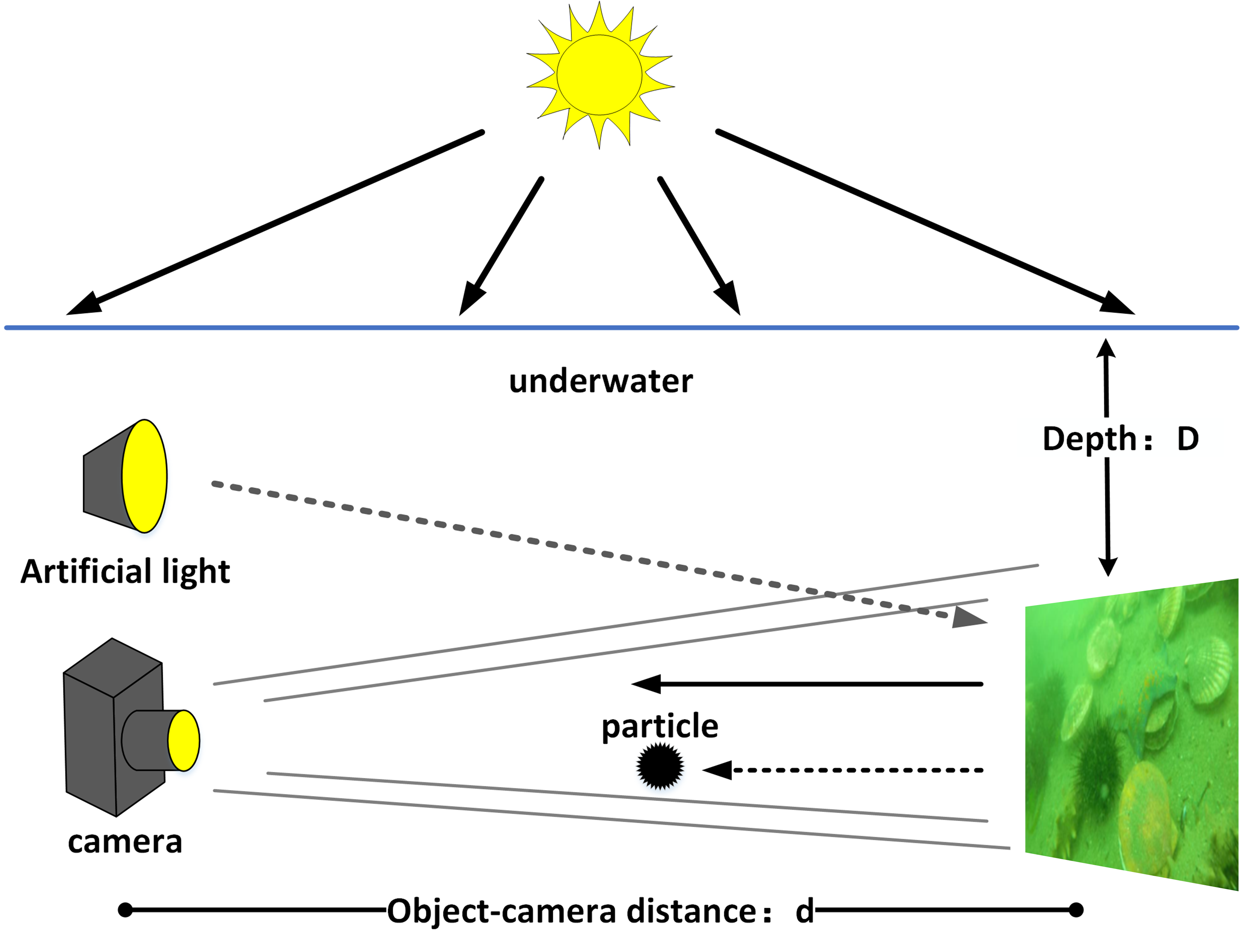}
\caption{The radiance perceived by the camera $I_\lambda$ is the sum of
 direct signal and background scattering. The black arrow represents the direct signal containing scene information, while the dashed arrow represents the scattered signal reflected by underwater suspended particles.}
\label{opmodel}
\end{figure}

To estimate background light, an efficient formula was proposed in \cite{zhao2015deriving}
as $B_\lambda =\kappa E_\lambda /c_\lambda$, where  $E_\lambda$ is the underwater illumination and $\kappa$ is a scalar defined by the camera system.
As illustrated in Fig.\ref{opmodel},  the underwater illumination can be simplified as the sum of incident light and artificial light  \cite{chiang2011underwater}:
\begin{equation}\label{illumination}
E_\lambda(x) = \omega_a E_\lambda^S e^{-c_\lambda D} +  \omega_b E_\lambda^{A} e^{-c_\lambda d(x)},
\end{equation}
where $\omega_a$ and $\omega_b$ are two weights, $E_\lambda^S$ is light on the water surface, $E_\lambda^{A}$ is artificial light, $D$ is the water depth and $d(x)$ is the scene distance from object to the camera.
Therefore, the background light $B_\lambda$ can be calculated  by the formula $B_\lambda(x) =\kappa E_\lambda(x) /c_\lambda$, where $\kappa$ is a scalar defined by the camera system.
Based on the camera's principle \cite{chiang2011underwater}, the expression $J_\lambda(x):=J_\lambda^{gt}(x) E_\lambda(x) / E_\lambda^S$ is an underwater image with new lighting condition $E_\lambda(x)$ for the in-air images $J_\lambda^{gt}$.
Therefore, the refined simulation formula can be expressed as follows:
\begin{equation}\label{uw model}
I_\lambda(x)=t_\lambda(x)J_\lambda^{gt}(x) \frac{E_\lambda(x)}{E_\lambda^S } + \frac{\kappa E_\lambda(x)}{c_\lambda} (1-t_\lambda(x)).
\end{equation}




\begin{table}[t]
  \centering
   \small
	\tabcolsep=3.5pt
	\renewcommand\arraystretch{1}
  \centering
  \caption{ The coefficients $e^{-c_\lambda}$ are employed to synthesize underwater images.} \label{attenuation coefficients}
    \begin{tabular}{c|c|c|c|c|c|c|c|c|c}
    \hline
\hline
       Types   & IA & IB   & II    & III  & 1 &  3 &5  &7 &9 \\
      \hline
         blue & 0.98& 0.97& 0.94& 0.89 & 0.88& 0.8&0.67 &0.5& 0.29  \\
         green &0.96& 0.95& 0.93& 0.89& 0.89& 0.82& 0.73 & 0.61&0.46  \\
         red  & 0.81& 0.83&  0.80 &0.75 & 0.75& 0.71&0.67 &0.62&0.55\\
          \hline
\hline
    \end{tabular}%
\end{table}%

\subsection{Parameter ranges in simulation formula}
Table \ref{notation} provides the parameter selection range in formula \eqref{uw model}, covering as many different underwater environments as possible.
To generate the synthetic underwater image dataset, we utilize the NYU-V1 dataset \cite{silberman2011indoor}, which consists of a total of 3733 RGB images and their corresponding depth maps.
As shown in table \ref{attenuation coefficients}, the Jerlov water types \cite{zaneveld1995light} cover the common attenuation coefficients $c_\lambda$
of seawater types.
Due to the complete absorption of light beyond 20 meters, this article considers water depths $D$ ranging from 5 meters to 20 meters.
Therefore, many underwater cameras are equipped with a high-brightness artificial light source.
This paper uses a two-dimensional Gaussian distribution with a beam pattern to simulate artificial light in water, which is given as follows
$E_\lambda^A = \mathcal{P}(\widetilde{x}|E_\lambda^{art}, \sigma).$
Here, $\widetilde{x}$ is a randomly selected light source from image, which
has the strongest artificial light.
Due to the high brightness values of artificial light sources, the range of the
peak value of  $E_\lambda^{art}$ is set to [0.7,1].
The range of values for the standard deviation $\sigma$ is [0.2, 1.1], which controls the illumination range of artificial light on the image.
Randomly selecting parameters during the training process helps the model adapt to a wide range of underwater environments, thereby enhancing its generalization ability.


\begin{table}[htbp]
 \footnotesize
	\tabcolsep=2pt
	\renewcommand\arraystretch{1}
  \centering
  \caption{These parameters are used to generate the synthetic underwater image dataset.}\label{notation}%
    \begin{tabular}{|c|c|c|}
 \hline
 Note & Description & Range\\
 \hline
 $D$ & Water depth & $[5m,20m]$\\
 $d$ &Transmission distance & NYU-V1 dataset \cite{silberman2011indoor} \\
 $c_\lambda$ & Attenuation coefficients & Table \ref{attenuation coefficients} \\
 $E_\lambda^S$ &Air light&   $[0.7,1]$\\
 $E_\lambda^{art}$ & Peak value of artificial light &  $[0.7,1]$\\
 $\widetilde{x}$ & Location of $E_\lambda^{art}$ & A random point in image\\
 $\sigma$ & Coverage of  artificial light & Random rate  $[0.2,1.1]$\\
 $\omega_a, \omega_b$ & Weights of lighting & $\omega_a\in[0,1] ~\mbox{and}~\omega_a+\omega_b=1$\\
 $\kappa$ &Camera system parameter & $[0.7,1.1]$\\
 \hline
 $E_\lambda^A$ &Artificial light& $E_\lambda^A=\mathcal{P}(\widetilde{x}|E_\lambda^{art}, \sigma)$ \\
 $E_\lambda$ & Underwater illumination & estimated by \eqref{illumination}\\
 $t_\lambda$ & Transmission map & $t_\lambda=e^{-c_\lambda d}$ \\
 $B_\lambda$ & Background light & $B_\lambda=\kappa E_\lambda /c_\lambda$\\
 \hline
    \end{tabular}%
\end{table}%

\begin{figure*}[t]
\centering
\includegraphics[width=0.5\textwidth]{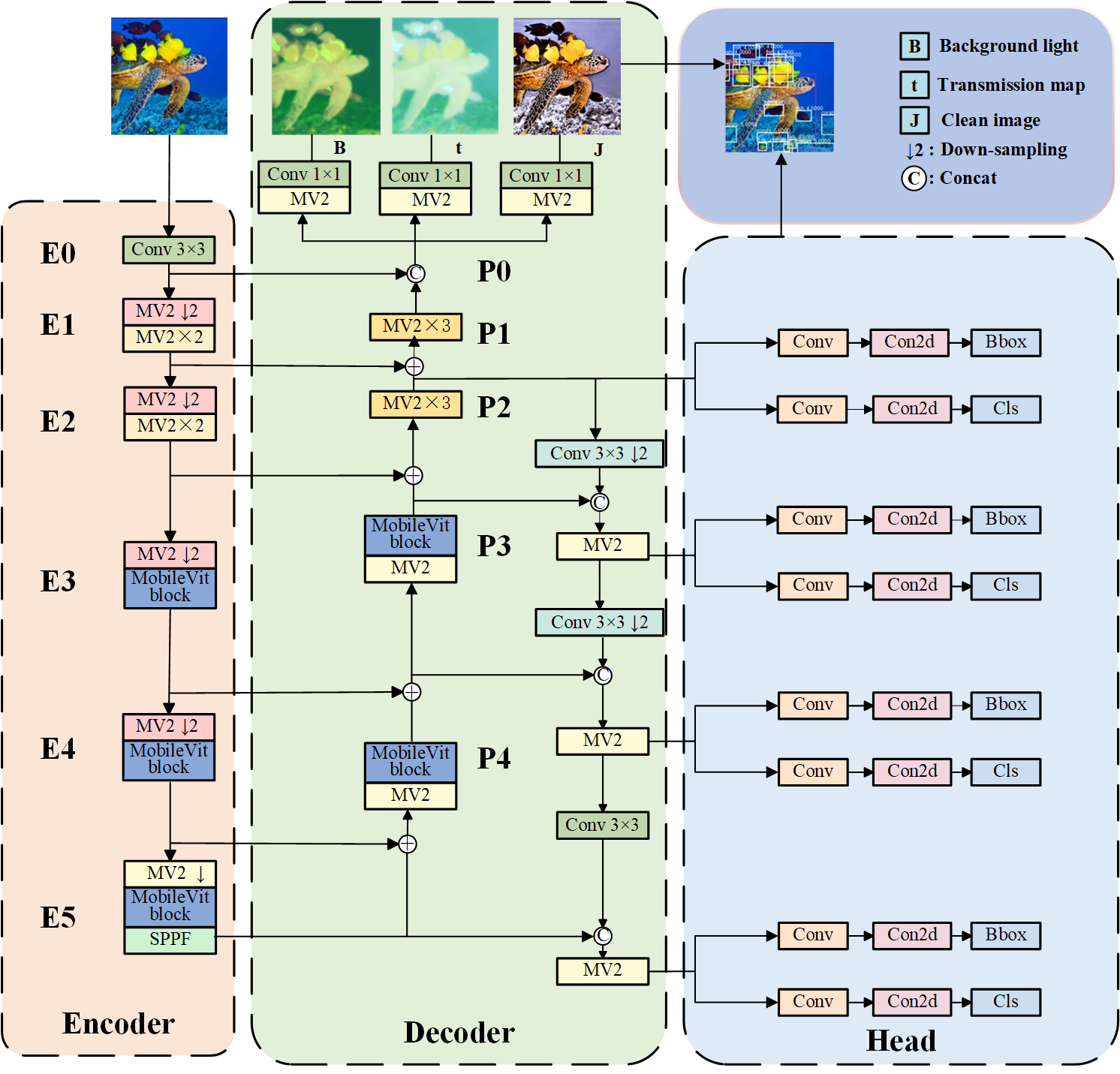}
\caption
{A lightweight model integrates image enhancement and object detection.
Here, the inverse residual structures are represented as MV2.
The image enhancement task divides an underwater image into a clean image, background light, and transmission maps, facilitating the self-supervised enhancement of real underwater images.
During the image enhancement decoding process, a path aggregation module is introduced to fuse multi-scale feature maps, and a decoupled anchor-free detection head is employed to identify underwater targets.}
\label{LUIEO}
\end{figure*}

\section{Network design}\label{sec4}

\subsection{Integrated Image Enhancement and Object Detection Model}
Fig.\ref{LUIEO} shows the proposed integrated structure of image enhancement and object detection,
consisting of three components: the encoder, decoder, and detection head.
The encoder extracts features from the original image, which are denoted as $\{E_0, E_1, E_2, E_3, E_4, E_5\}$.
The decoder structure enhances the underwater image through upsampling
and establishes lateral connections with the feature layers of the encoder
to generate a set of multi-scale fusion features $\{P_0, P_1, P_2, P_3, P_4\}$.
Then, three branch networks decode the feature $P_0$ and output the clean image, background light, and transmission maps.
These physical variables enable the enhancement model to perform self-supervised training on underwater images.
To detect underwater targets, a path aggregation module is used to fuse multi-scale feature maps $\{E_5, P_4, P_3, P_2\}$,
obtaining fused multi-scale features $\{D_1, D_2, D_3, D_4\}$ to detect objects of various sizes.
Subsequently, the decoupled anchor-free detection heads are employed to detect targets in these feature maps, using two separate convolutions
for classification and regression to output the category and bounding box position independently.

\subsection{Lightweight network module}
This paper uses lightweight components to construct the network structure, mainly containing inverted residual module and MobileNetV3.
These modules are designed to extract features efficiently with a lightweight structure, reducing computational complexity
and memory requirements, and making the network more suitable for mobile devices.

\begin{figure*}[h]
\centering
\includegraphics[width=0.5\textwidth]{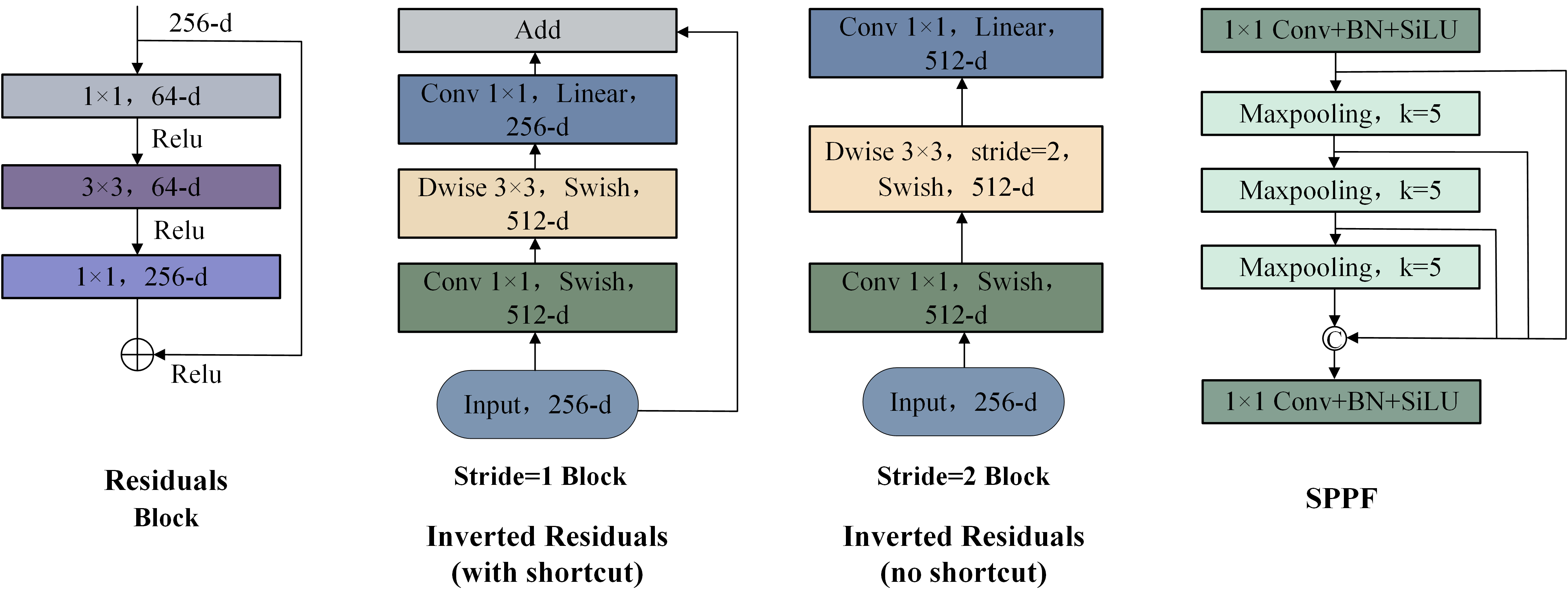}
\captionsetup{justification=centering}
\caption
{Illustration of residual network structure, inverted residual network structure, and spatial pyramid pooling fast (SPPF) structure.}
\label{block}
\end{figure*}

\textbf{Inverted Residuals:}
The inverted residual structure aims to extract underwater image features with a lightweight structure and is utilized for downsampling and feature extraction.
The residual is used to connect the input and output during feature extraction to avoid gradient divergence.
The process of inverted residual structure involves dimensionality expansion, convolution, and dimensionality reduction,
which is the reverse of the residual structure process.
Therefore, it is named inverse residual structure.
As shown in Fig.\ref{block}, the inverted residual structure uses depthwise separable convolutions to extract features,
thereby reducing computational complexity while maintaining high accuracy.
Since the features extracted by depthwise separable convolutions are dependent on the input feature dimensions,
the inverted residual structure initially employs a $1\times 1$ convolution to increase the dimensionality.
The activation function represents the complex relationship between the input and output of a neural network,
influencing the performance of deep learning models.
The Swish activation function is utilized in the inverted residual structure,
which can produce large gradients during forward propagation to alleviate the problem of gradient vanishing.
Finally, a linear activation function replaces the Swish function when reducing dimensions.
The reason behind this is that the activation function  Swish will set negative features to zero, resulting in a loss of some information.

\textbf{MobileViT-V3:}
Convolutional neural networks and vision transformers are commonly used deep learning models for image processing.
A key difference between them is the prior assumption of image data.
CNNs assume local connectivity and translation invariance of features,
enabling them to establish local information dependencies.
In contrast, the self-attention layer in Vision transformers can capture global receptive fields and establish comprehensive global dependencies.
However, transformers exhibit a lack of local correlation and translation invariance,
which requires sufficient training data to achieve better performance.

\begin{figure}[!htbp]
\centering
\includegraphics[width=0.4\textwidth]{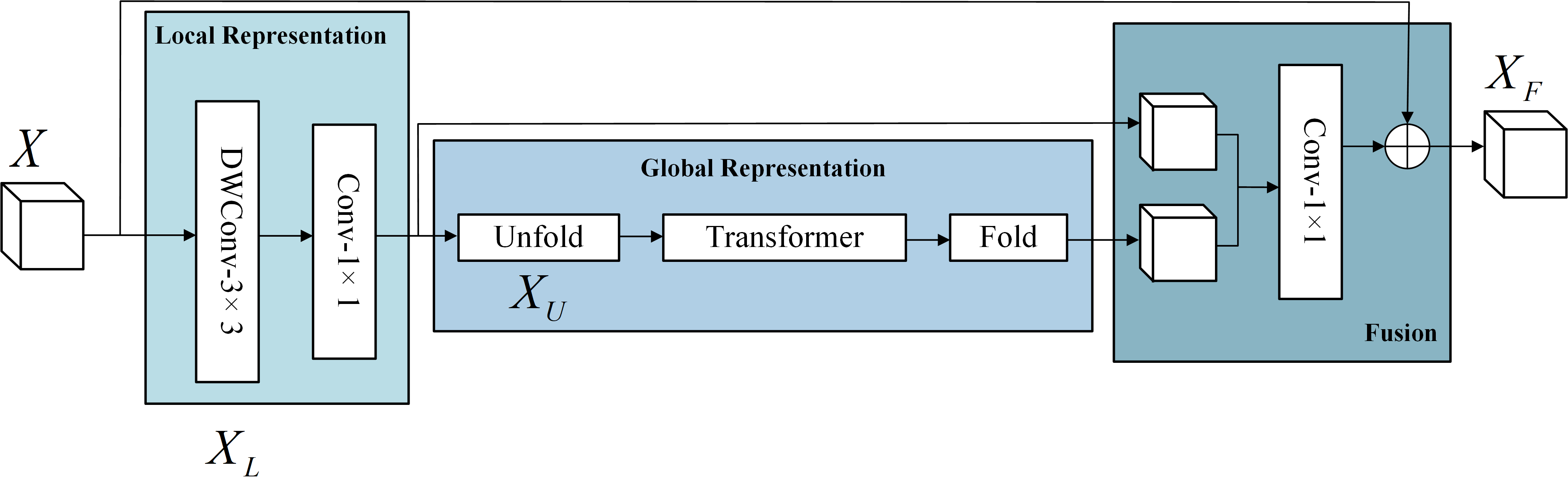}
\caption{Illustration of MobileViT architecture combining CNN with Transformer.}
\label{mobilevit}
\end{figure}

MobileViT combines the advantages of both standard convolutional and transformer architectures,
allowing it to effectively learn both local and global information with a relatively small number of model parameters.
As illustrated in Fig.\ref{mobilevit}, MobileViT employs a $3 \times 3$ depthwise separable convolutional
layer to encode the input tensor $X \in \mathbb{R}^{H \times W \times C}$,
where $H$, $W$, and $C$ represent the height, width, and number of feature channels of the feature maps, respectively.
Subsequently, point-wise convolutions are employed to project the local spatial features into high-dimensional
spatial features $X_L \in \mathbb{R}^{H \times W \times d}$, where $d$ is the spatial dimension.

To enable MobileViT to learn a global representation with spatial inductive bias,
the feature layer $X_L$ is divided into non-overlapping patches $X_U \in \mathbb{R}^{P \times N \times d}$,
where $P = wh$, $N = HW/P$ is the number of patches,
and $h, w$ represent the height and width of each patch (set to $h = w = 2$ in this paper).
These non-overlapping feature maps are subsequently processed through $L$ stacked transformers to extract global information from $X_U$.
The self-attention in MobileViT focuses on the relationships between the patches, rather than on individual pixels within the patches.
This allows the model to attend to the global context while reducing the computational burden that would otherwise arise from attending to all pixels in the image.
Due to the redundancy of information present in adjacent areas of the image, MobileViT can effectively reduce the high computational burden caused by transformers.
Compared with ViT, it loses the spatial arrangement of pixels,
MobileViT preserves the order of patches and the spatial locations of pixels within each patch,
which can fold the tensor to obtain features consistent with the dimensions of the local feature layer $X_L$.
Subsequently, a $1 \times 1$ convolution is applied to adjust the channels of the feature maps,
and the global information is fused with the local information.
Finally, the residual connection is established between the input features and the fused features
to optimize the deep network in the architecture.

\textbf{Physical module:}
To use a physical model for supervised training, this paper uses residual connections to map the feature layer $P_0$
to clean images, background light, and transmission maps, respectively.
As shown in Fig.\ref{block}, the residual connection includes an inverse residual module
to fuse multi-scale information and a $1 \times 1$ convolution to reduce dimensionality.
The simulated underwater image guides these three network structures to decompose the underwater image into three physical variables, providing prior knowledge and physical constraints for real underwater images.
Therefore, physical constraints allow for simultaneous training of image enhancement and object detection tasks.


\textbf{SPPF:}
SPPF is processed through three $5 \times 5$ pooling layers to extract feature maps of different sizes,
which enhances the model's ability to detect objects of varying sizes.

\textbf{Detect head:}
As shown in Fig.\ref{LUIEO}, the anchor-free detection heads sequentially predict the centre point and object class
in the multi-scale feature maps $\{D_1, D_2, D_3, D_4\}$.
The detection head utilizes a structure that decouples classification and detection to focus on their respective tasks
and improve performance.
Each branch contains two convolutional blocks and a separate Conv2d layer for boundary prediction and class prediction.
The regression task has 4 feature channels for predicting the positions of the left, right, top, and bottom sides of the bounding box.
The classification branch predicts object types in each bounding box, with feature channels matching the number of categories.

\section{Loss function}
Our proposed integrated network aims to achieve both high-quality visual performance and precise detection results.
Therefore, the loss function considers both sub-tasks to effectively guide the optimization process of multi-task joint learning.

\subsection{Image enhancement loss}
\textbf{Clean image loss.}
The $L_{J}$ is used to supervise the loss between the predicted and clean images:
\[L_{J}=\|J-J^{gt}\|_1.\]

\textbf{Background light image loss.}
Due to the issues of the small number of pixels occupied and severe background light attenuation for distant objects,
it is difficult to accurately estimate the intensity of distant light, leading to significant errors.
Therefore, the loss in logarithmic space is used instead of the $L_1$ loss to suppress the impact of inaccurate long-distance estimation and make the network focus on nearby information.
Due to the influence of light absorption, there is a significant difference in the values of the three channels of background light.
As a result, the loss function of background light is as follows:
\[L_{back}=\sum_{\lambda\in \{R,G,B\}}( \|ln(B_\lambda-B_\lambda^{gt})\|_1+1) ,\]
where $B_\lambda^{gt}=\kappa E_\lambda/c_\lambda$ can be regarded as the ground truth of the background light.


According to physical formulas, the  background light is closely related to the scene depth.
Inspired by monocular scene depth estimation \cite{hu2019revisiting}, this paper adopts gradient loss and normal loss to overcome boundary distortion and distortion problems.
Therefore, the Sobel operator is used to extract the gradient between the
background light and the ground truth image as follows:
\[L_{grad} =\sum_{\lambda\in \{R,G,B\}} ( \|ln( \nabla(B-B^{gt})) \|_1+1)).\]
Here, the $\nabla(B-B^{gt})$ denotes the extraction of gradient information using the Sobel operator.

The normal vector error on the surface of the objects is used to learn the subtle variations in light intensity, which is perpendicular to the gradient direction.
The normal vector loss is defined as:
\[L_{normal} = \sum_{\lambda\in \{R,G,B\}} \big( 1-\frac{<n^B,n^{B^{gt}}>}{\sqrt{<n^B,n^{B}>}\sqrt{<n^{B^{gt}},n^{B^{gt}}>}}\big). \]
where $n^B$ is the normal vector of background light $B$.
Therefore, the total loss of background light is defined as:
\[L_{B} = L_{back} + L_{grad} + L_{normal}.\]

\textbf{Transmission map loss.}
Similar to the background light, the loss of transmission map $L_t$ is similar to that of background light, where the reference transmission map is given by $t^{gt}_\lambda= e^{-c_\lambda d}$.

\textbf{Physical model loss.}
Based on the physical model \eqref{image model}, the underwater image can be calculated by:
\[\widetilde{I}_\lambda(x)= J_\lambda t_\lambda + B_\lambda(1-t_\lambda).\]

Consequently, the loss function based on the physical process is defined as follows:
\[L_{I} =\|I_\lambda  -\widetilde{I}_\lambda\|_1. \]

\textbf{Image enhancement loss.}

 For image enhancement task,  the loss function for training simulation images combines the learning of three physical variables and physical constraints, which can be written as
\begin{equation}\label{enhance function}
L_{enhance}=L_J+L_B+L_t+ c_{I}L_{I}.
\end{equation}
 For training real underwater images, the loss function is $L_{enhance}=L_{I}$.



\subsection{Object Detection loss}

The anchor-free detection head directly predicts the position and class of the target on the feature map, without relying on predefined anchor boxes.
Therefore, this paper uses YOLOv8's \cite{li2024efficient} classification loss and regression loss as the loss functions for object detection.
Despite the advantage of fast convergence, the decoupled structure leads to misalignment between classification and regression tasks.
Therefore, task alignment learning techniques \cite{akkaynak2019sea} are used to align classification prediction and regression tasks,
where the degree of alignment is defined as follows:
\[t = s^\alpha \times u^\beta .\]
Here, $s$ is the predicted class score, $u$ is the intersection over union (IoU) value between the predicted box and the ground truth box, $\alpha$ and $\beta$ are weights. In the paper,  the hyperparameter settings follow those of YOLOv8, which are $\alpha=0.5$ and $\beta=6$.
Thus, $t$ can achieve task alignment between classification and regression through classification scores and IoU optimization, directing the network to focus on high-quality prediction frames during training.

\textbf{Classification loss.}
The predicted category scores are represented as $p=(p_1,\cdots,p_i,\cdots,p_{na})\in R^{bs\times na \times cls_{num}}$, while the corresponding learning labels are denoted by $y\in R^{bs\times na \times cls_{num}}$.
Here, $bs,na,cls_{num}$  are denoted as batch size, anchor number, and number of target categories, respectively.
Therefore, the classification loss is calculated using the binary cross-entropy loss function:
\[L_{cls}(y,p) = \sum_{i=0}^{na}(-y_i log(\sigma(p_i)) - (1-y_i)log(\sigma(p_i)) ) / \sum_{i}^{na} y_i,\]
where $\sigma(p_i) = 1/(1+exp(p_i))$.

\textbf{Regression loss.}
Intersection over Union (IoU) is a metric used to describe the overlap between two bounding boxes.
In regression tasks, the ratio between the target box and the predicted box is used to measure the degree of regression of a box.
The CIoU loss extends the IoU loss by incorporating aspect ratio and center distance, improving the fit between the predicted $b\in R^{bs\times na \times 4}$ and target boxes $b^{gt}\in R^{bs\times na \times 4}$ by considering overlap area, center point distance, and aspect ratio:
\[L_{CIoU}(b,b^{gt}) = 1-IoU(b,b^{gt}) + \frac{\rho^2(b,b^{gt})}{c^2}+\alpha \mu(b,b^{gt}), \]
which $1-IoU$ represents the loss of intersection over union.
Here, $\rho^2(b,b^{gt})/c^2 $ and $\mu(b,b^{gt})$ respectively represent the loss of center point distance and aspect ratio of the predicted boxes and the label boxes.
The $\alpha$ is used to  adjust the loss of center point distance and aspect ratio, which is set to $1$ as shown in YOLOv8\cite{li2024efficient}.

\textbf{Object detection loss.}
Therefore, the loss function of the object detection task is a weighted sum of the classification loss and the localization loss:
\begin{equation}\label{object loss}
L_{obj} = w_{cls}L_{cls} +w_{box} L_{CIoU},
\end{equation}
where $w_{cls}=0.5$ and $w_{box}=7.5$. The parameter settings follow the configuration of the YOLOv8.
\subsection{Integrated loss function for image enhancement and object detection}
Finally, the overall loss function in this paper is a weighted combination of the image enhancement loss and the object detection loss:

\begin{equation}\label{total loss}
L =\alpha L_{enhance} + (1-\alpha)L_{obj}.
\end{equation}
The  hyper-parameter $\alpha$ is used to adjust the importance of two learning tasks.
For real underwater images, image enhancement and object detection tasks are equally important, which requires adjusting hyper-parameter $\alpha$ to achieve this goal.
The simulated underwater images only serve the image enhancement task, guiding the model to learn prior knowledge and physical processes. Therefore, the parameters are set to $w_{enhance} = 1,  w_{obj}=0$.

\begin{figure*}[b]
\centering
\includegraphics[width=0.85\textwidth]{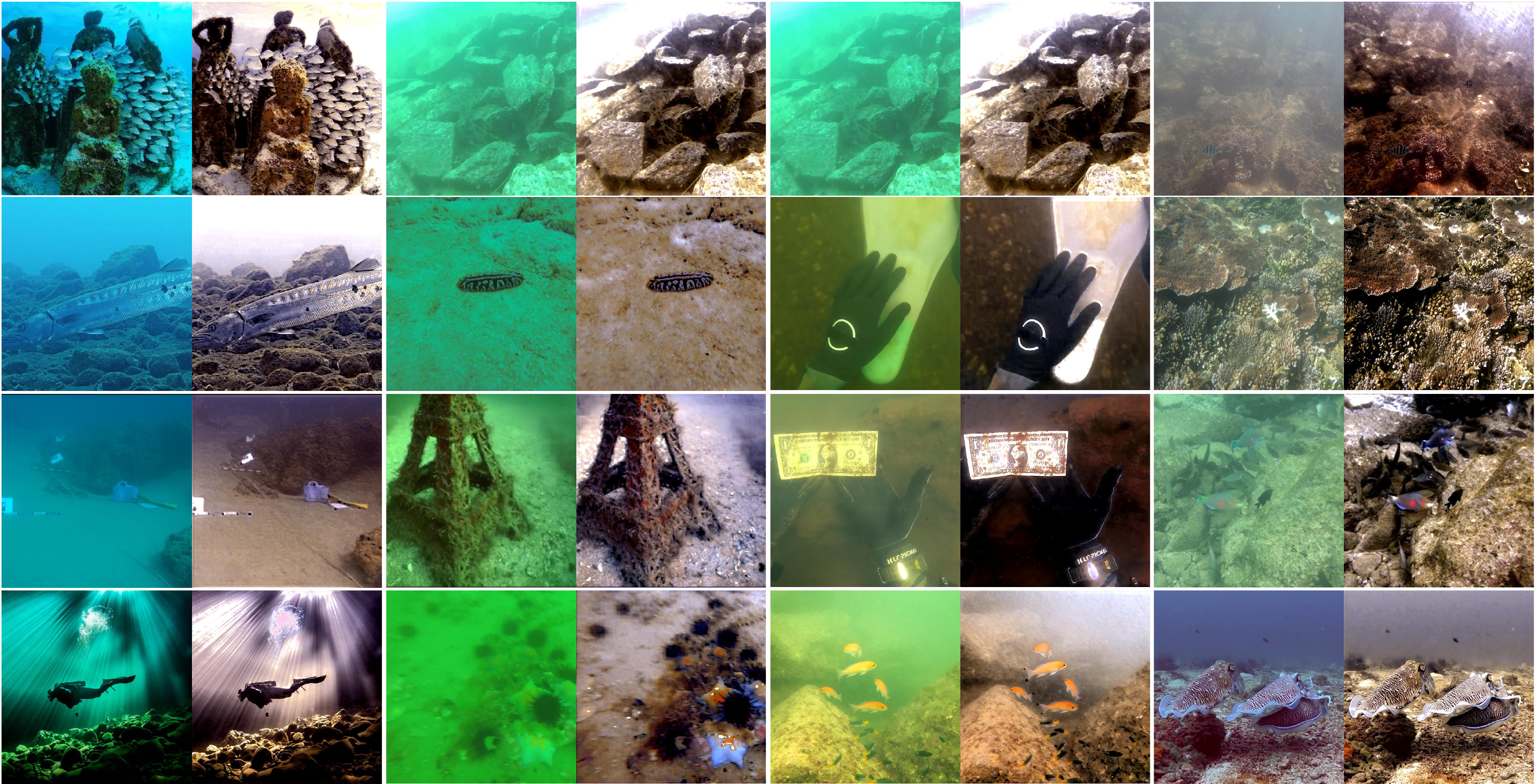}
\captionsetup{justification=centering}
\caption
{The enhanced results of proposed model for common underwater degradation types.}
\label{enhance degrade}
\end{figure*}

%
%
%

\begin{figure}[!htbp]
\centering
\includegraphics[width=0.5\textwidth]{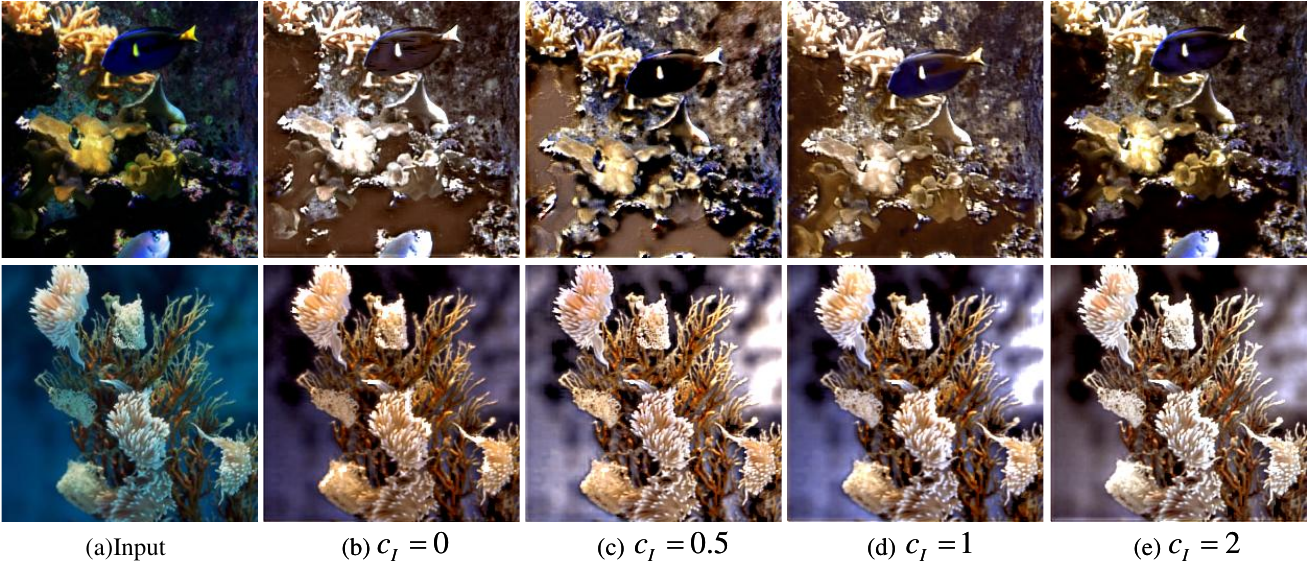}
\caption{The enhanced results of different values for hyper-parameters $c_I$.}
\label{cI}
\end{figure}

\section{Experiment and analysis}\label{sec5}
This section introduces the datasets used for image enhancement and object detection, along with implementation details and evaluation metrics to ensure the accuracy and reproducibility of the experiments.
We evaluate the enhancement performance of the model on underwater images with multiple degradation factors, demonstrating that the enhanced results  outperform existing methods.
In section \ref{detect compared}, the proposed model is compared with the existing underwater object detection methods,
and the experimental results show that the proposed approach excels in both inference speed and detection accuracy on multiple datasets.
Due to the simultaneous optimization of image enhancement and object detection tasks, it is easier to directly determine the detection results from the enhanced images.

\subsection{Datasets description and experimental metrics for image enhancement}

\textbf{UIEB\cite{li2019underwater}}is a dataset containing 950 images of different underwater environments, which includes 890 paired reference images. Here $200$ samples were selected to test the image
enhancement performance of the proposed model.

\textbf{U45\cite{li2019fusion} and UCCS\cite{duarte2016dataset} } are  underwater test datasets designed to evaluate the performance of different algorithms under common underwater degradations, such as color distortion, low contrast, and haze effects. All samples were selected  to test the enhancement effect of the model in different underwater environments.

The performance of our method is compared against four state-of-the-art underwater image enhancement methods:
Ucolor \cite{li2021underwater}, TACL \cite{liu2022twin}, U-Cycle \cite{yan2023uw} and TUDA \cite{wang2023domain}.

Due to the lack of paired clean images in underwater images,
UCIQE \cite{yang2015underwater} and UIQM \cite{panetta2015human}, two non-reference evaluation metrics, are used to evaluate the performance of enhanced images. A higher UCIQE or UIQM score indicates better human visual perception.

\begin{figure*}[b]
\centering
\includegraphics[width=0.9\textwidth]{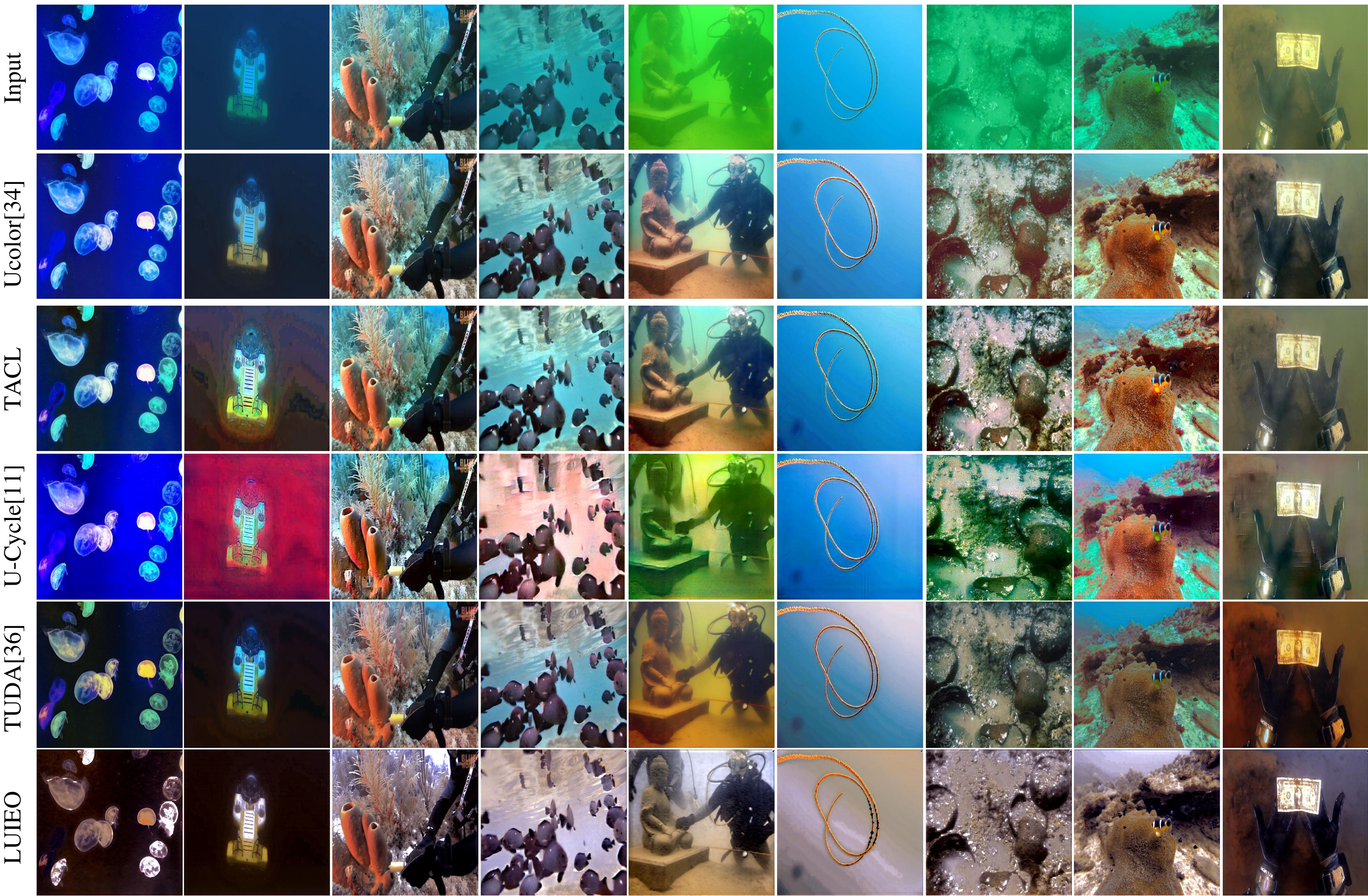}
\captionsetup{justification=centering}
\caption
{The visual comparison among the compared methods on tested datasets. The underwater images are listed in the first row, rows 2-4 show the results of the comparison methods, and ours are in the last row.}
\label{compared1}
\end{figure*}

\subsection{Datasets description and experimental metrics for object detection}
\textbf{RUOD\cite{fu2023rethinking}} consists of 14,000 underwater images,  annotated with 10 common aquatic organisms: holothurian, echinus, scallop, starfish, fish, corals, diver, cuttlefish, turtles, and jellyfish.
This dataset includes a wide variety of marine objects and diverse degradation factors, such as haze effects, color cast, and light interference.
All $4200$ validation samples from RUOD are used for evaluation, providing a large number of samples to ensure the generalization of the results.

\textbf{DUO\cite{liu2021dataset}} contains 7782 underwater images, which are used to detect four types of underwater organisms: sea cucumber, echina, scallop, and starfish.
Here, $1111$ samples were selected from the validation set to test the proposed model.

The evaluation metrics for object detection include precision, recall, mAP50, mAP50-95$^c$, and FPS.
Precision assesses the reliability of the model's predictions, indicating the proportion of predicted positive results that correspond to actual existing objects.
Recall measures the proportion of correctly identified objects relative to the total number of objects, representing the model's ability to detect all real objects without omissions.
The mean Average Precision (mAP50) comprehensively considers the recall and precision of the model, quantifying the detection accuracy at an IoU threshold of $0.5$.
It is a widely used metric for evaluating the overall performance of object detection models.
Here,  mAP50-95$^c$ represents the average map value for IoU thresholds of $0.5$, $0.75$, and $0.95$.
Frames per second (FPS) measures the number of frames processed per second, reflecting the running speed of the model.

\begin{figure*}[!htbp]
\centering
\includegraphics[width=0.9\textwidth]{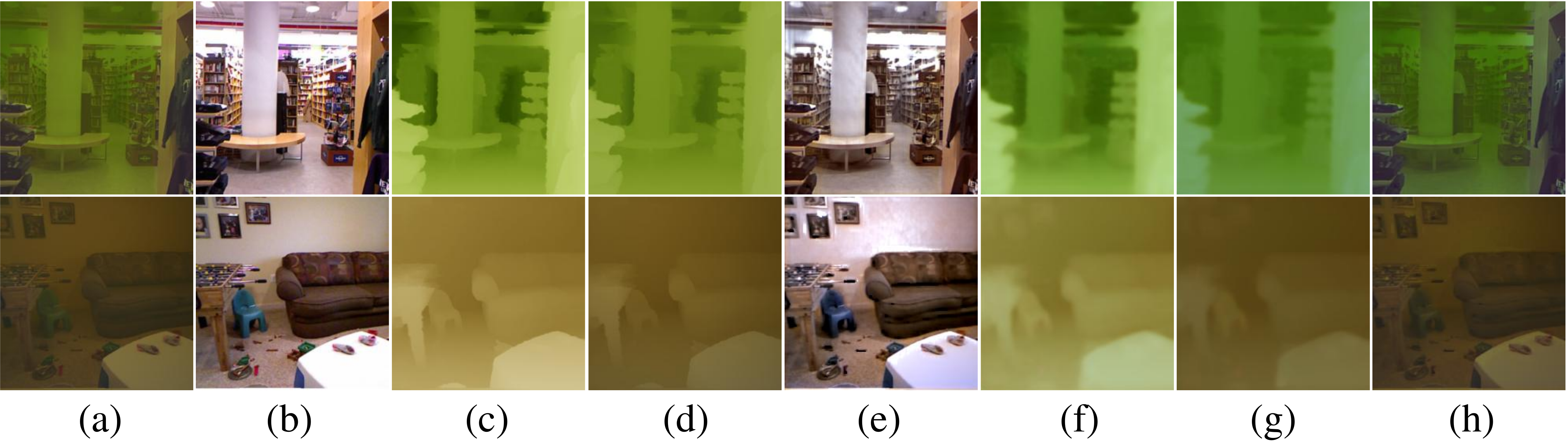}
\captionsetup{justification=centering}
\caption
{Results of estimated clean images, transmission maps and background lights, where (a) simulated underwater images, (b) true clean images, (c) the reference images of background light, (d)the reference images of transmission map, (e-f) are the corresponding prediction results, and (f) is the underwater images calculated by the predicted three variables.}
\label{JBT}
\end{figure*}

\subsection{Implementation details}
The lightweight model for the image enhancement task was trained on the NYU dataset, which contains $3799$ pairs of clean images and corresponding scene depths.
The object detection task was trained on the RUOD dataset, which provides $9800$ and $4200$ samples for training and validation, respectively.
The proposed network was trained on GeForce RTX $4090$ GPUs using PyTorch for $500$ epochs, alternating between simulated and real data training.
To address memory limitations, we used gradient accumulation to optimize the model, accumulating gradients over 5 steps with a batch size of 2 per iteration.
Multi-scale training, commonly used to improve model performance, was applied in this work.
A multi-scale sampler was used to collect data, with image sizes ranging from $256\times 256$ to $640\times 640$, along with random cropping and flipping for data enhancement.
The learning rate followed a cosine schedule with a warmup, starting from $0.0001$ and increasing to $0.001$ over $5000$ warmup iterations.
Both the training and evaluation were conducted on GeForce RTX 4090 GPUs.

\begin{figure*}[!htbp]
\centering
\includegraphics[width=0.7\textwidth]{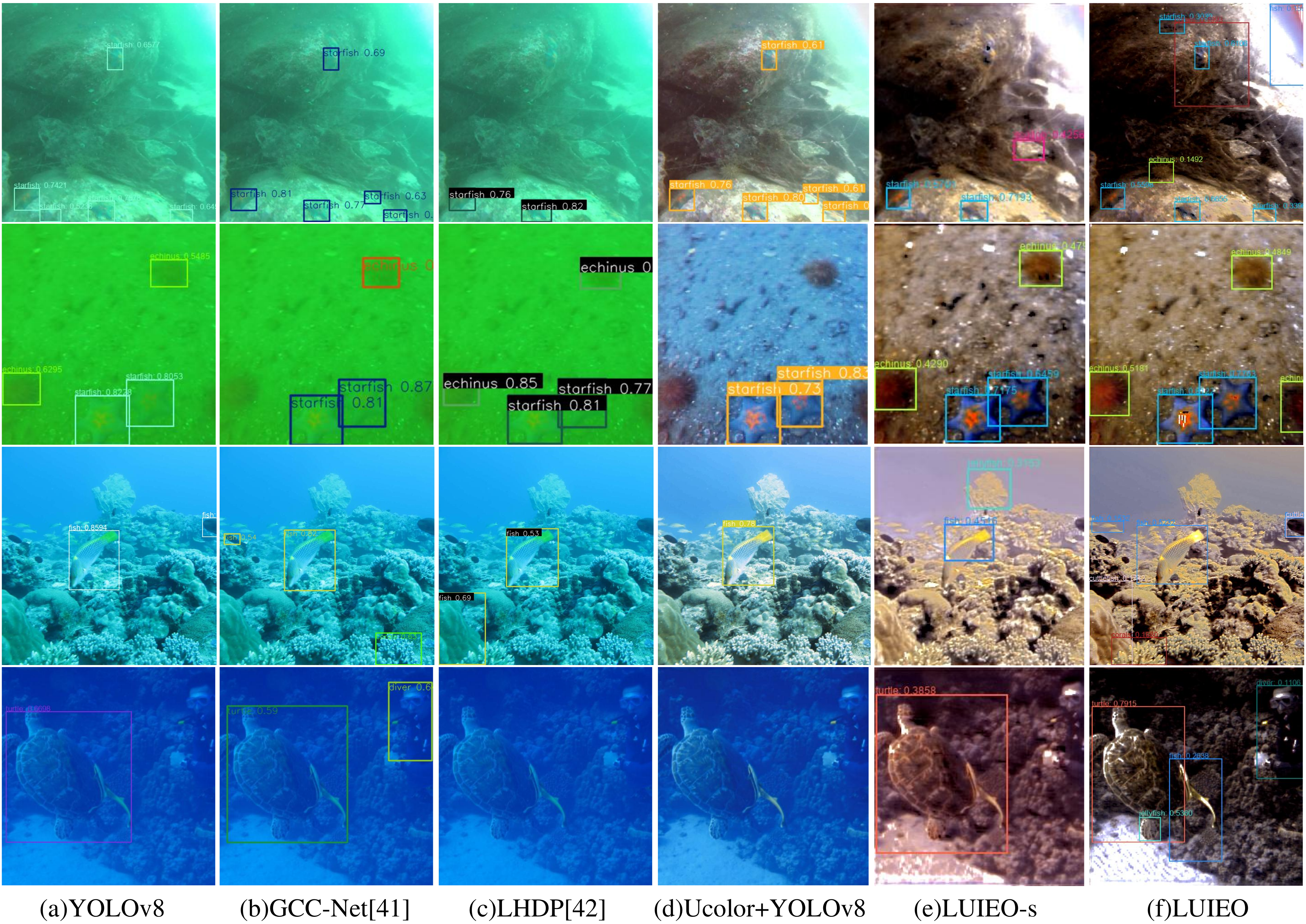}
\captionsetup{justification=centering}
\caption
{Comparison results between the proposed object detection model and the comparison methods on  typical underwater degraded images.
}
\label{compared obj1}
\end{figure*}

\subsection{Hyper-parameter selection of loss functions }
For the proposed multi-task network, we first adjust the hyper-parameters of each sub-task loss functions to ensure that the network optimizes in the correct direction.
Subsequently, we adjust the importance of the two tasks through the $\alpha$ of \eqref{total loss}.
For object detection, we follow the setting of  hyper-parameters in YOLOv8, which allows the network to be optimized without the need for adjustment.
Therefore, the method proposed in this article only requires adjusting two hyper-parameters $c_I$ of \eqref{enhance function} and $\alpha$ in \eqref{total loss}.

\begin{table}[htbp]
  \centering
  \caption{Analysis results of hyperparameter selection $c_I$ for  loss function $L_{enhance}$.}
    \begin{tabular}{|c|cccc|}
    \hline
          \multicolumn{1}{|c|}{$c_I$} & 0 & 0.5 &1 &2 \\
              \hline
     UIQM         &   4.5072 &  \textbf{4.7000} &  4.5789   &  4.5705 \\
      UCIQE    & 0.5075      &  \textbf{0.5084}  & 0.4987   & 0.5051\\
    \hline
    \end{tabular}%
  \label{loss enhance}%
\end{table}%

To roughly find feasible parameter $c_I$, we set a group parameters and conducted $10$ epochs for image enhancement training.
Table \ref{loss enhance} shows results of UIQM and UCIQE on 100 randomly selected underwater images with different parameters $c_I$.
The visual performances are shown in Fig. \ref{cI}.
Therefore, based on UIQM and visual performances, we selected  parameters  $c_I=0.5$ as weights for the loss function.

\begin{figure}[!htbp]
\centering
\includegraphics[width=0.3\textwidth]{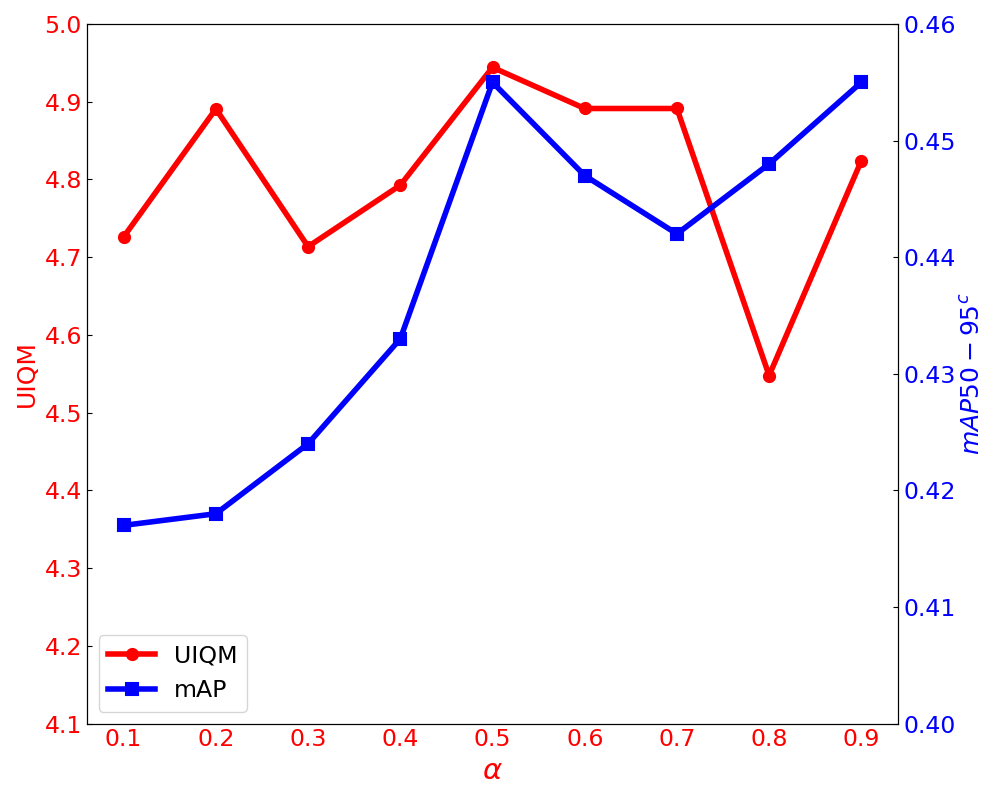}
\caption{Analysis results of hyper-parameter selection $\alpha$ for total loss function \eqref{total loss}.}
\label{alpha_UIQM}
\end{figure}

In this paper, the two sub-tasks are equally important.
Therefore, we conducted experiments using different values $\alpha$ in loss \eqref{total loss}, and each experiment was trained for 10 epochs.
As shown in Fig. \ref{alpha_UIQM}, the evaluation metrics for the two subtasks indicate that they are of equal importance when $\alpha$ is set to $0.5$.

Therefore, through this selection method, we determined the hyperparameters used in the model presented in this paper.

\subsection{Evaluation of Image Enhancement Models}

The ability to enhance various types of degraded underwater images is an important capability of image enhancement models.
As shown in Fig.\ref{enhance degrade}, we illustrate the enhancement performance of the model on images affected by blur, low light, and color degradation, demonstrating the effectiveness of the model.
It can be observed that the proposed model can generalize to enhance various types of underwater images, effectively removing the visual interference of degradation factors.
To further demonstrate the effectiveness of the proposed model, Fig.\ref{compared1} presents a visual comparison with several enhancement methods, including Ucolor \cite{li2021underwater}, TACL \cite{liu2022twin}, U-Cycle \cite{yan2023uw}, and TUDA \cite{wang2023domain}.
Compared to the performance of other methods, the proposed model effectively enhances underwater images with multiple degradation types.
However, other methods struggle to generalize and enhance images with multiple types of degradation simultaneously.
Specifically, Ucolor \cite{li2021underwater} and TACL \cite{liu2022twin} exhibit difficulties in enhancing bluish-degraded images, whereas U-Cycle \cite{yan2023uw} and TUDA \cite{wang2023domain} encounter challenges with green-biased images.
In comparison, the proposed model generalizes well on various degradation types and produces enhanced images that align with human visual perception, which can be attributed to the use of simulated prior information and physical constraints.

To quantitatively demonstrate the generalization of the image enhancement model, Table \ref{Pre-training-uciqe} shows the averaged UCIQE and UIQM metrics of the comparison method on multiple datasets.
Here, the UCIQE and UIQM of the underwater images are used as the baseline to illustrate the performance of the enhancement method.
It can be observed that different methods have higher evaluation metrics than the baseline on multiple datasets, indicating that image enhancement helps improve visual performance.
Among these methods, the performance of the method proposed in this article is higher than that of the comparison method, which is consistent with the visual effect shown in Fig.\ref{enhance degrade}.
Therefore, the proposed method has better generalization ability than the comparative method in diverse degraded underwater environments.

\begin{table}[htbp]
  \centering
  \tabcolsep=1.5pt
\renewcommand\arraystretch{1}
  \caption{ This table shows the UIQM and UCIQE of the proposed method and the comparative method on multiple underwater datasets.  }
    \begin{tabular}{|c|cc|cc|cc|}
    \hline
  \multirow{2}{*}{\diagbox[width=1.7cm]{$\mathrm{Methods}$}{$\mathrm{Datasets}$}}  & \multicolumn{2}{c|}{UIEB} & \multicolumn{2}{c|}{U45} & \multicolumn{2}{c|}{UCCS}  \\
  \cline{2-7}
      &  \multicolumn{1}{c}{UIQM}  & \multicolumn{1}{c|}{UCIQE} & \multicolumn{1}{c}{UIQM}  & \multicolumn{1}{c|}{UCIQE}& \multicolumn{1}{c}{UIQM}  & \multicolumn{1}{c|}{UCIQE}\\
          \hline
    Baseline&2.6847 &0.3875&1.6717 &0.3526  &2.0221 & 0.3701\\ 
          \hline
    Ucolor \cite{li2021underwater} &3.3620& 0.3980  &3.0810& 0.4024  &3.3271& 0.3971   \\
    TACL \cite{liu2022twin} &4.5654& 0.4619& 4.1353& 0.4053 &4.5172& 0.4159\\ 
    U-Cycle \cite{yan2023uw} &4.2872& 0.4545  &3.6640& 0.4672&4.2116& 0.4841 \\
    TUDA \cite{wang2023domain}&4.5504& 0.5157&4.1879& 0.4062 &4.3341& 0.4333\\
    \hline
    LUIEO &\textbf{4.7997}& \textbf{0.5384} &\textbf{4.6841}& \textbf{0.5248}&\textbf{4.5397}& \textbf{0.5164}\\
    \hline
    \end{tabular}%
  \label{Pre-training-uciqe}%
\end{table}%

To verify the accuracy of the estimated physical variables, we compared the estimated background light and transmission map on the simulated underwater images with the reference images.
As shown in Fig. \ref{JBT}, we can observe that our estimations are visually similar to the reference background light and transmission map, which confirms the effectiveness the proposed method.
In addition, the underwater images calculated by the predicted three variables are similar to the degraded images,  indicating the effectiveness of physical constraints.

\subsection{Performance of object detection models}\label{detect compared}

\begin{figure*}[!htbp]
\centering
\includegraphics[width=0.85\textwidth]{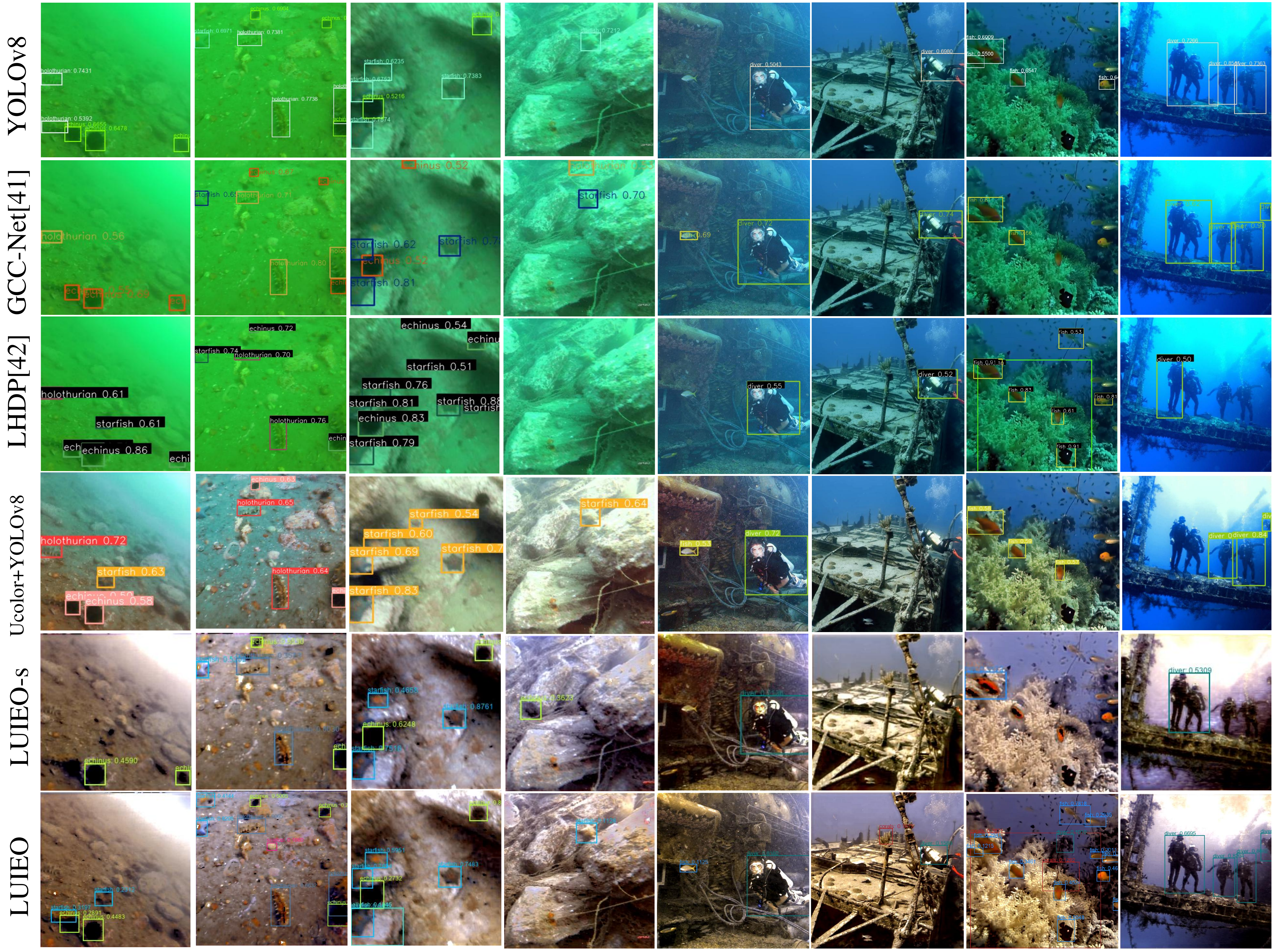}
\captionsetup{justification=centering}
\caption
{This figure provides a visual comparison of the object detection results between the proposed LUIEO model and the comparative models across various types of degraded images. The first three rows display the results of the comparative experiments, while the last row presents the detection results obtained with the proposed model.
}
\label{compared obj2}
\end{figure*}

To comprehensively compare the two tasks,  the compared methods includes three methods of first enhancing and then detecting, namely Ucolor+YOLOv8, LUIEO+YOLOv8, and a model that separates the two tasks (denoted as LUIEO-S).
Here, LUIEO+YOLOv8 indicates that LUIEO only performs the image enhancement task,
and the enhanced images are used to train YOLOv8.
In addition, the comparison methods also include three methods that solely focus on object detection, namely YOLOv8\cite{li2024efficient}, GCC-Net\cite{dai2024gated} and LHDP\cite{fu2023learning}, all of which have been retrained  on the RUOD dataset.

As shown in table \ref{map},
the detection accuracy of GCC-Net\cite{dai2024gated} and LHDP\cite{fu2023learning} methods designed  specifically  for underwater detection is higher than that of the baseline YOLOv8, indicating the effectiveness of these methods.
Compared to the GCC-Net\cite{dai2024gated} and LHDP\cite{fu2023learning}, combining image enhancement with the target detection task is more effective in improving detection accuracy.
We observed that the detection accuracy of LUIEO+YOLOv8 is higher than that of Ucolor+YOLOv8.
This is mainly attributed to the generalization ability of the proposed method on various degraded underwater images.
In addition, the detection accuracy of LUIEO+YOLOv8 is similar to that of LUIEO-S,
which demonstrates the effectiveness of the target detection network designed in this paper.
Finally, we compared the two separate task LUIEO-S with the integrated model LUIEO.
It can be observed that LUIEO outperformed LUIEO-S in both detection accuracy and efficiency, indicating that information exchange in multi-task learning can promote the learning of sub-tasks.
Additionally, we compare the inference speed of our method against the contrastive methods under identical conditions.
As shown in Table \ref{map}, the lightweight structure enables the proposed model to achieve $80$ FPS, outperforming the speed of the comparison methods and meeting the requirements for real-time processing.

Fig.\ref{compared obj1} presents a visual comparison of the proposed detection method with the compared method on severely degraded underwater images, including color degradation, bright light interference, and low illumination.
In the first row of images, the degradation factors include color decay and strong light interference, which make it difficult for YOLOv8 to accurately detect objects.
Although the detection results of  GCC-Net\cite{dai2024gated} and LHDP\cite{fu2023learning}  outperform YOLOv8, accurate assessment of their performance remains challenging on degraded images.
The integrated model in this paper displays detection results on the enhanced image with improved visual performance, facilitating easy verification of object detection in the images.
Fig.\ref{compared obj2} further illustrates the detection results of the proposed method and comparison methods on underwater images with various types of degradation.
It can be observed that the proposed model demonstrates high accuracy in detecting underwater targets.

Therefore, through a comprehensive comparison, this fully demonstrates the advantages of the integrated model proposed in this paper in terms of detection efficiency, accuracy and visual effects.

\begin{table*}[htbp]
  \centering
  \caption{This table presents the results for precision, recall, mAP50, mAP50-95$^c$, and FPS of the proposed model and other methods on the datasets RUOD and DUO. The best results here are highlighted in bold.}
    \begin{tabular}{|c|cccc|cccc|c|}
    \hline
    Dataset & \multicolumn{4}{c|}{RUOD: $4200$ test images}& \multicolumn{4}{c|}{DUO: $1000$ test images} &\\
    \hline
    Method & \multicolumn{1}{l}{Precision} & \multicolumn{1}{l}{Recall} & \multicolumn{1}{l}{mAP50} & \multicolumn{1}{l|}{mAP50-95$^c$}  & \multicolumn{1}{l}{Precision} & \multicolumn{1}{l}{Recall} & \multicolumn{1}{l}{mAP50} & \multicolumn{1}{l|}{mAP50-95$^c$} & \multicolumn{1}{c|}{FPS} \\
    \hline
    YOLOv8   & 0.793   & 0.575   & 0.698   & 0.473  & 0.753   & 0.535   & 0.654   & 0.351  & 120.48 \\
    GCC-Net\cite{dai2024gated}  & 0.771   & 0.542   & 0.662   & 0.365  & 0.812   & 0.597   & 0.710   & 0.487   & 39.65 \\
LHDP\cite{fu2023learning}    & 0.785   & 0.552   & 0.675  & 0.372   & 0.827   & 0.587   & 0.727   & 0.493  & 60.31 \\
\hline
Ucolor+YOLOv8 &  0.835    & 0.609    & 0.731 & 0.499&  0.794    & 0.562    & 0.681 & 0.388 & 2.17 \\
LUIEO+YOLOv8  &  0.838    & 0.611    & 0.742 & 0.503 &  0.799    & 0.565    & 0.692 & 0.391& 66.35 \\
LUIEO-S       &  0.833    & 0.605   & 0.729 & 0.492 &  0.790    & 0.560    & 0.679 & 0.383& 36.33 \\
\hline
LUIEO     & \textbf{0.841}    & \textbf{0.614}   & \textbf{0.755}   &  \textbf{0.506}  & \textbf{0.829}    & \textbf{0.604}   & \textbf{0.695}   &  \textbf{0.397} & 80.56 \\
    \hline
    \end{tabular}%
  \label{map}%
\end{table*}%

\begin{figure}[!htbp]
\centering
\includegraphics[width=0.45\textwidth]{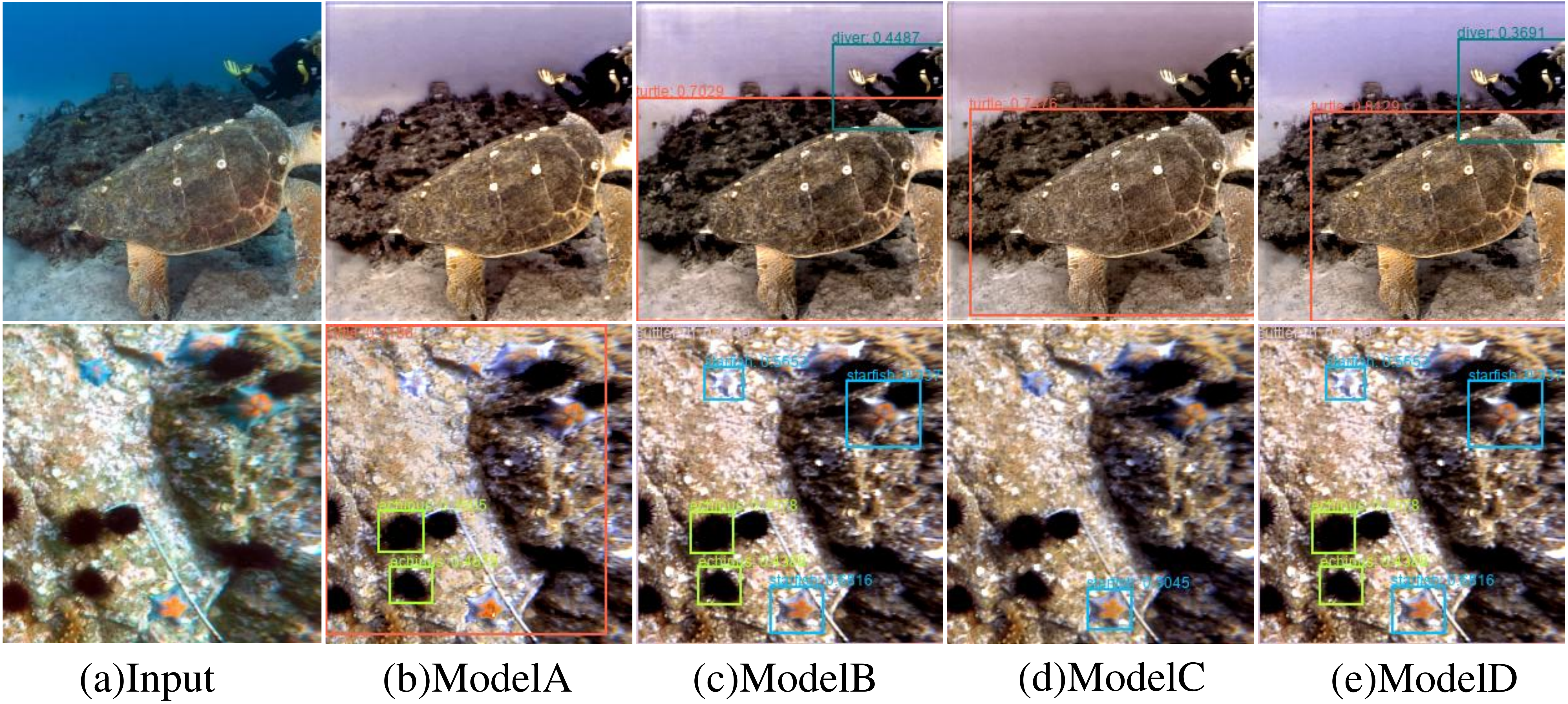}
\caption{Visual comparison of individual SPPF and mobileViT on model contributions. The final column displays the enhancement and detection results of the model.}
\label{ablation model}
\end{figure}

\begin{figure}[!htbp]
\centering
\includegraphics[width=0.45\textwidth]{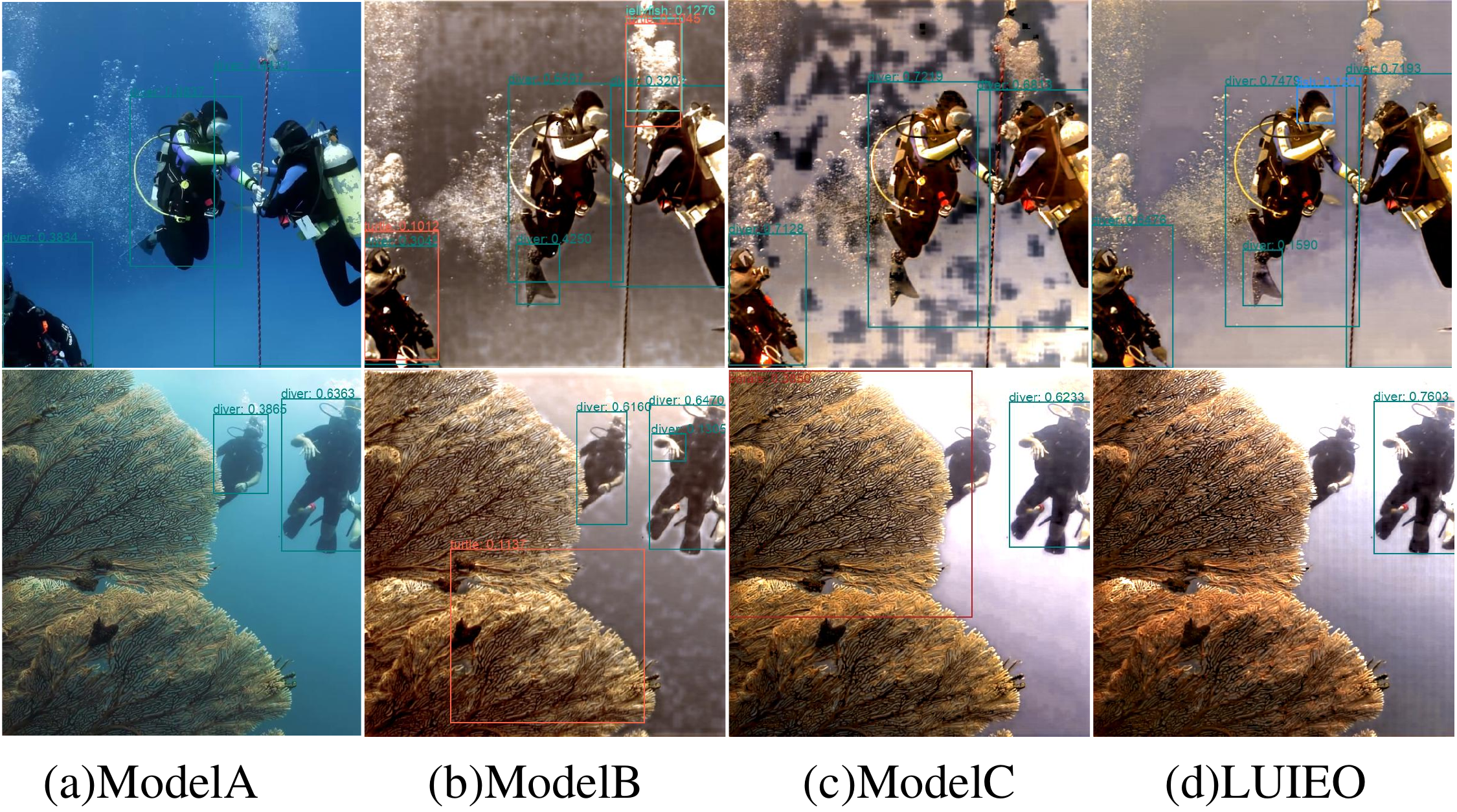}
\caption{Visual comparison of simulation priors and physical constraints on model contributions. The final column displays the enhancement and detection results of the model.}
\label{ablation obj1}
\end{figure}

\subsection{Evaluation of the model's complexity}
The evaluation metrics for network complexity usually include the number of parameters (Params(M)), model size (Size(M)), and floating point operations (FLOPs(G)), which showcase the model's complexity, storage requirements, and computational demands.
Table \ref{model complex} shows the model complexity evaluation metrics of
the compared methods and the proposed method, using an input size of $256\times256\times3$ for comparisons.
It can be observed that the  three complexity metrics of our proposed method are superior to most of the comparative methods.
This is primarily attributable to the integrated network design and lightweight components.
Therefore, our method has the potential to be deployed on underwater computing platforms with limited computing resources.

\begin{table}[htbp]
  \centering
  \caption{Comparison results of model complexity quantitative metrics.
  The best results here are highlighted in bold, while the second best ones are underlined.}
  
    \begin{tabular}{|l|rrr|}
    \hline
    Methods & \multicolumn{1}{l}{Params(M)} & \multicolumn{1}{l}{Size(M)} & \multicolumn{1}{l|}{FLOPs(G)} \\
    \hline
    TACL  &   11.86    & 199.00  &56.86  \\
    U-Cycle &  8.92   &  165.60  &37.92\\
    TUDA  &  4.28  & 48.80  & 26.34 \\
    \hline
    YOLOv8 &  \textbf{3.01}  &  \textbf{5.94}  & \textbf{0.66}  \\
    GCC-Net&  38.31   & 146.13  & 21.57 \\
    LHDP &   75.59  & 288.34  & 15.40 \\
    \hline
    LUIEO &   \underline{4.09}    & \underline{33.80}     & \underline{5.22}\\
    \hline
    \end{tabular}%
  \label{model complex}%
\end{table}%

\subsection{Ablation study}

\textbf{Individual effect of  component SPPF and MobileViT: }
In the following, we designed ablation studies to validate the impact of SPPF and MobileViT components on the model.
In table \ref{sppf vit ablation}, the symbol $\times$ indicates the absence of this component.
Here, the modelB means removing the attention structure and replacing it with an inverted residual structure.
Compared to ModelA, the  ModelB introduces the SPPF module, which significantly improves the detection accuracy.
The ModelC introduces attention mechanism MobileViT, which significantly improves the image enhancement and detection accuracy of the model.
Therefore, by introducing attention mechanism and SPPF, our model can effectively accomplish these two tasks, as shown in table \ref{sppf vit ablation}. The visualization results in Fig. \ref{ablation model} also confirm this conclusion.

\begin{table}[htbp]
  \centering
  \caption{The ablation studies for the components of SPPF and MobileViT.}
    \begin{tabular}{|c|cc|cc|c|cc|c|}
    \hline
    \multicolumn{1}{|c}{\multirow{2}[0]{*}{Models}} & \multicolumn{2}{|c|}{Components} &  \multicolumn{3}{c|}{RUOD} \\
          \cline{2-6}
&SPPF&ViT& \multicolumn{1}{l}{UCIQE} & \multicolumn{1}{l|}{UIQM} & \multicolumn{1}{l|}{mAP50}   \\
    \hline
 ModelA  &$\times$   & $\times$ &  0.4713     &  3.9826     &  0.611      \\
 ModelB   & $\checkmark$ & $\times$ &    0.4764   &  4.1322     &  0.663      \\
 ModelC   &$\times$  & $\checkmark$ &   0.5128    &  4.4845     &  0.701      \\
 LUIEO    & $\checkmark$  & $\checkmark$& \textbf{0.5364}   & \textbf{4.7388}  & \textbf{0.756}  \\
 \hline
   \multicolumn{1}{|c}{\multirow{2}[0]{*}{Models}}& \multicolumn{2}{|c|}{Components} &  \multicolumn{3}{c|}{DUO} \\
          \cline{2-6}
&SPPF&ViT& \multicolumn{1}{l}{UCIQE} & \multicolumn{1}{l|}{UIQM} & \multicolumn{1}{l|}{mAP50}   \\
 \hline
  ModelA  &$\times$   & $\times$   &  0.4328     &   3.8846    & 0.514 \\
 ModelB   & $\checkmark$ & $\times$    &    0.4422   &  3.9214     & 0.599 \\
 ModelC   &$\times$  & $\checkmark$   &   0.4655    &  4.2739   & 0.654 \\
 LUIEO    & $\checkmark$  & $\checkmark$& \textbf{0.5064} & \textbf{4.5287}  & \textbf{0.695} \\
    \hline
    \end{tabular}%
  \label{sppf vit ablation}%
\end{table}%

 \textbf{Effects of prior information and physical constraints on object detection task}
Due to the lack of supervised information in underwater images, this paper introduces prior information and physical constraints from simulated images to train an integrated model for image enhancement and object detection.
Therefore, we designed ablation experiments to verify the contributions of simulation prior information and physical model constraints to object detection.
Specifically, the modelA indicates that the model is without both the physical and simulation processes, while modelB and modelC represent the introduction of physical model constraints and the simulation process, respectively.

As shown in Fig.\ref{ablation obj1}, the  modelA results in the model performing only the object detection task, with the detection results shown in the underwater images.
Although modelB can enhance underwater images, its enhancement effects are limited.
This is because it lacks prior knowledge of simulation, making it difficult to rely solely on physical information to predict clean images, background light, and transmission maps.
The modelC fails to consistently enhance images with various types of degradation.
The lack of physical information and reliance solely on simulation prior knowledge leads to unstable image enhancement.
Therefore, the proposed LUIEO model incorporates simulation as prior knowledge and physical constraints, enabling the model to generalize and enhance various types of degradation.

\begin{table}[htbp]
  \centering
  \caption{This table presents the object detection results of the ablation experiments. The modelA means that the model without both the physical and simulation processes, while modelB and modelC represent the introduction of physical model constraints and the simulation process, respectively.}
    \begin{tabular}{|c|c|c|c|c|c|}
    \hline
    \multicolumn{1}{|c}{Models} & \multicolumn{1}{|c|}{sim} & \multicolumn{1}{c|}{phy}& \multicolumn{1}{c|}{P} & \multicolumn{1}{c|}{R} & \multicolumn{1}{c|}{mAP50}  \\
    \hline
    ModelA & $\times$ & $\times$ & 0.751 & 0.378 & 0.512  \\
    ModelB      & $\times$ & $\checkmark$ &  0.771 & 0.398 & 0.532 \\
    ModelC      & $\checkmark$ & $\times$ & 0.854 & 0.491 & 0.723  \\
    LUIEO     & $\checkmark$ & $\checkmark$ & \textbf{0.871} & \textbf{0.604} & \textbf{0.755} \\
    \hline
    \end{tabular}%
  \label{ablation}%
\end{table}%

To more accurately demonstrate its effectiveness, table \ref{ablation} presents the object detection results of the ablation experiments.
The results show that the detection accuracy of modelA demonstrates that the designed network structure can perform the object detection task.
While physical constraints alone can improve detection accuracy to some extent, their effect is limited because it lacks simulation prior information.
The inclusion of simulation prior information allows the model to effectively perform both image enhancement and object detection tasks, achieving better results than modelB and modelA.
After adding physical constraints, the model proposed in this paper can utilize the prior information to self-supervise underwater images, thus improving both image enhancement and object detection performance.
Consequently, the use of simulation information and physical constraints is effective for model training.

\section{Conclusion}\label{sec6}

This paper proposes a lightweight underwater object detection method integrating image enhancement and object detection into a unified framework, aiming to improve object detection accuracy and achieve visually appealing results.
The refined simulation formulation provides valuable prior information for the image enhancement task, allowing the model to generalize well across various types of degraded images.
This enables object detection models to utilize enhanced feature maps to improve detection accuracy and facilitate intuitive evaluation of detection performance.
The enhanced images are used to display the results of object detection,  facilitating intuitive observation and evaluation of detection performance.
 However, the simulated images cannot be fully approximated to real underwater images, and there is a domain gap between them.
Therefore, in future work, we consider the fusion of optical images and sonar to obtain accurate underwater scene information, which will contribute to tasks such as underwater scene restoration and 3D object detection, without considering the domain gap between  synthetic images.

\section*{Acknowledgments}
The work was supported by the National Natural Science Foundation of China (NSFC 12401703) and   Natural Science Foundation of Shanxi Province, China(202403021212256).

\section*{Disclosures}
The authors declare no conflicts of interest.


\bibliographystyle{IEEEtran}
\bibliography{sample}
 \vspace{-5mm}
\begin{IEEEbiography}[{\includegraphics[width=0.95in,height=1.25in,clip]{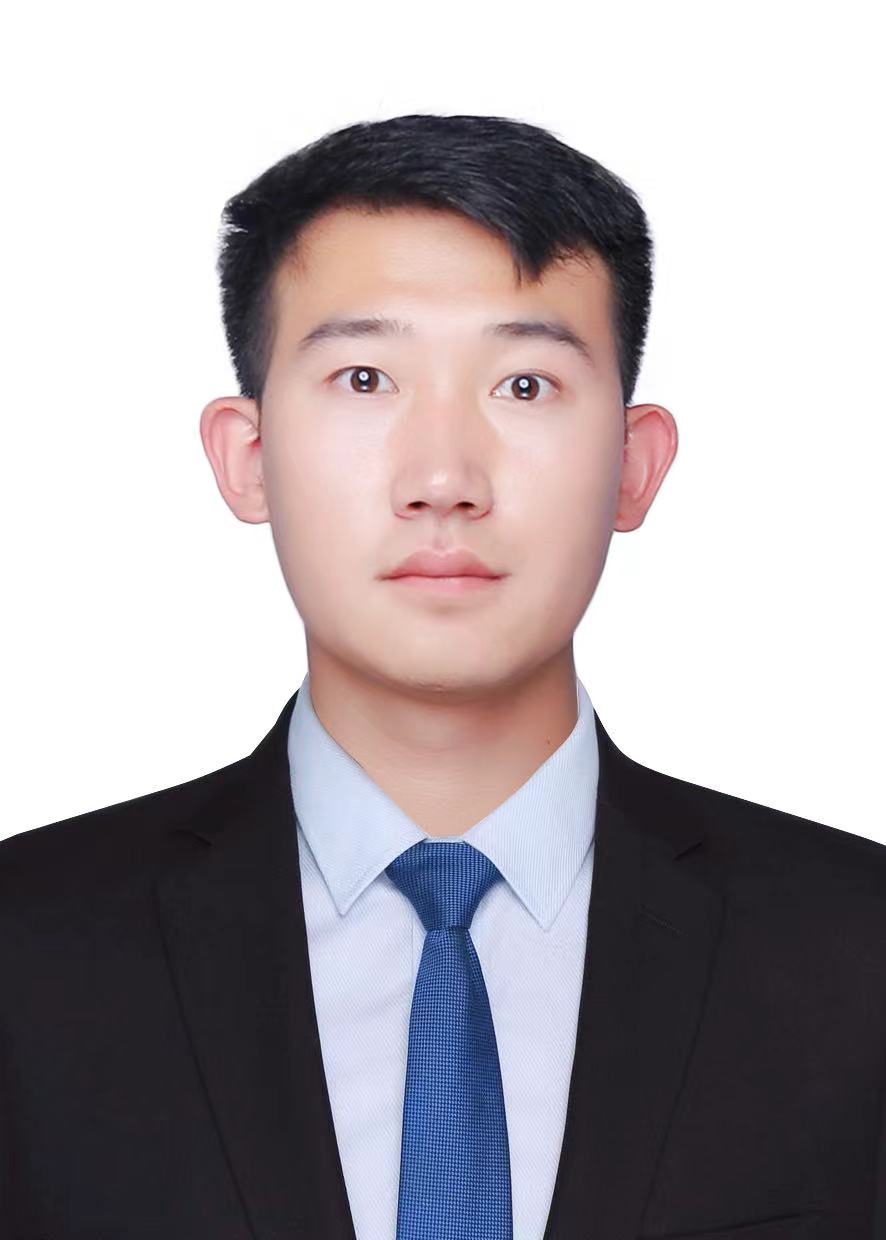}}]{Bin Li}
received the B.S. degree in school of mathematics and statistics from Datong University, China, in 2022.
He is currently pursuing the M.S. degree in mathematics with the North University of China, Taiyuan, China.
His research interests include object detection and compute vision.
\end{IEEEbiography}
 \vspace{-5mm}
\begin{IEEEbiography}[{\includegraphics[width=0.95in,height=1.25in,clip]{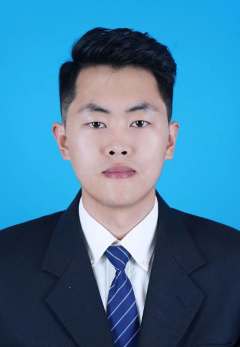}}]{Zhenwei Zhang}
received the Ph.D. degree in mathematics from Tianjin University, Tianjin, China, in 2023.
He is currently working with the school of Mathematics, North University of China, Taiyuan, China. His research interests include computer vision and image processing.
\end{IEEEbiography}
 \vspace{-5mm}
\begin{IEEEbiography}[{\includegraphics[width=0.95in,height=1.25in,clip]{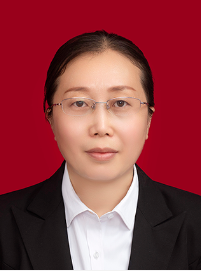}}]{Li Li}
received the Ph.D. degree from North University of China, Taiyuan, China, in 2013.
She is currently a Professor at the School of Computer and Information Technology, Shanxi University, Taiyuan, China. Her research interests include complex network dynamics and vegetation pattern dynamics.
\end{IEEEbiography}
  \vspace{-5mm}
\begin{IEEEbiography}[{\includegraphics[width=0.95in,height=1.25in,clip]{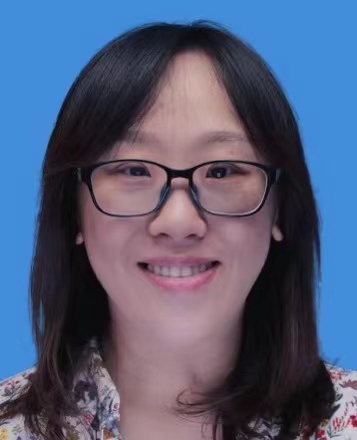}}]{Yuping Duan}
received the Ph.D degree from Nanyang Technological University, Singapore, in 2012.
She is currently a Professor with the School of Mathematical Sciences, Beijing Normal University, Beijing, China. Her research interests include numerical optimization, computer vision, and image processing.
\end{IEEEbiography}

\vfill

\end{document}